\documentclass[final]{cvpr}

\usepackage{times}
\usepackage{epsfig}
\usepackage{graphicx}
\usepackage{caption}
\usepackage{amssymb}
\usepackage{amsmath,amsfonts,amsthm,bm}
\usepackage{setspace}

\newcommand{\Name}{DriveGAN}

\usepackage{array}
\newcolumntype{P}[1]{>{\centering\arraybackslash}p{#1}}

\usepackage[pagebackref=true,breaklinks=true,colorlinks,bookmarks=false]{hyperref}

\newcommand{\ours}{\text{DriveGAN}}

\def\cvprPaperID{4843} 
\def\confYear{CVPR 2021}

\begin{document}

\title{DriveGAN: Towards a Controllable High-Quality Neural Simulation}

\author{\vspace{1mm}Seung Wook Kim$^{1,2,3}$
\hspace{0.5cm}
Jonah Philion$^{1,2,3}$
\hspace{0.5cm}
Antonio Torralba$^{4}$
\hspace{0.5cm}
Sanja Fidler$^{1,2,3}$
\\
$^1$NVIDIA 
\hspace{2em} 
$^2$University of Toronto 
\hspace{2em} 
$^3$Vector Institute  
\hspace{2em} 
$^4$ MIT
\\
{\tt\small \{seungwookk,jphilion,sfidler\}@nvidia.com}
\hspace{2em}
{\tt\small torralba@mit.edu}
}

\maketitle

\begin{abstract}
    Realistic simulators are critical for training and verifying 
    robotics systems. 
    While most of the contemporary simulators are hand-crafted, a scaleable way to build simulators is to use machine learning to learn how the environment behaves in response to an action, directly from data.
    In this work, we aim to learn to simulate a dynamic environment directly in pixel-space, by watching unannotated sequences of frames and their associated actions. 
   We introduce a novel high-quality neural simulator referred to as {\ours}  that achieves controllability by disentangling different components without supervision.
    In addition to steering controls, it also includes controls for sampling features of a scene, such as the weather as well as the location of non-player objects.
    Since {\ours} is a fully differentiable simulator, it further allows for re-simulation of a given video sequence, offering an agent to drive through a recorded scene again, possibly taking different actions.
    We train {\ours} on multiple datasets, including 160 hours of real-world driving data.
    We showcase that our approach 
    greatly surpasses the performance of previous data-driven simulators, and allows for new key features not explored before.

\end{abstract}


\vspace{-4mm}
\section{Introduction}

The ability to \textit{simulate} is a key component of intelligence. 
Consider how animals make thousands of decisions each day.
Some of the decisions are critical for survival, such as deciding to step away from an approaching car.
Mentally simulating the future given the current situation is key in planning successfully.
In robotic applications such as autonomous driving, simulation is also a scaleable, robust and safe way of testing self-driving vehicles in safety-critical scenarios before deploying them in the real world. Simulation further allows for a fair comparison of different autonomous driving systems since one has control over the repeatability of the scenarios.

Desired properties of a good robotic simulator include accepting an action from an agent and generating a plausible next world state, allowing for user control over the scene elements, and the ability to re-simulate an observed scenario with plausible variations. This is no easy feat as the world is incredibly rich in situations one can encounter. Most of the existing simulators~\cite{Dosovitskiy17,VirtualHome2018,THOR,AirSim} are hand-designed in a game engine, which involves significant effort in content creation, and designing complex behavior models to control non-player objects. 
Grand Theft Auto, one of the most realistic driving games to date, set in a virtual replica of Los Angeles, took several years to create and involved hundreds of artists and engineers.  In this paper, we advocate for data-driven simulation as a way to achieve scaleability.





\begin{figure}[t!]
        \vspace{-1mm}
    \begin{center}
        \includegraphics[width=0.8\linewidth]{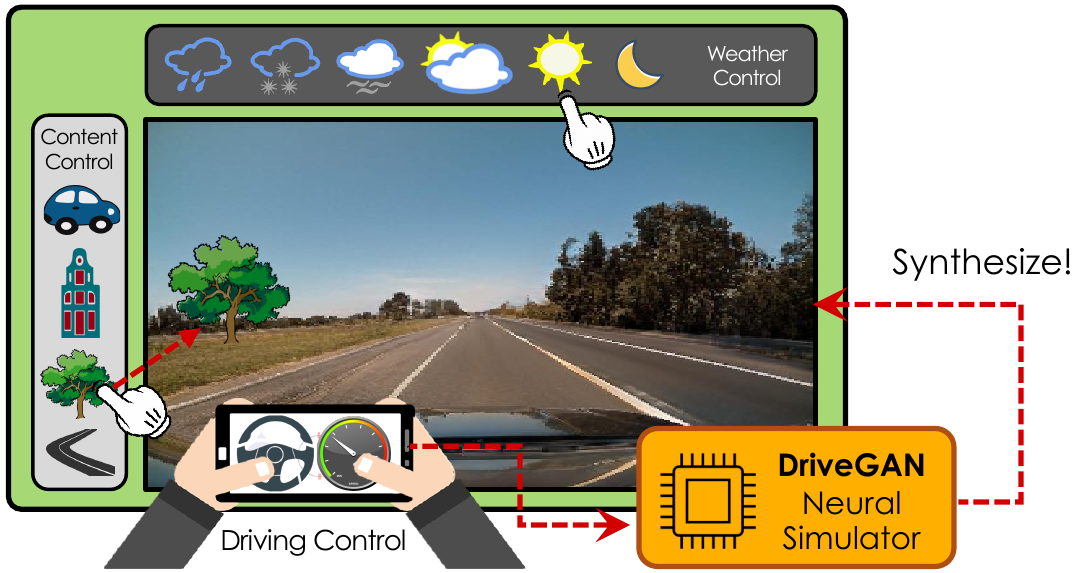}
    \end{center}
    \vspace{-4.5mm}
    \caption[]{
        \small We aim to learn a controllable neural simulator that can generate high-fidelity real-world scenes.
        {\Name} takes user controls (\eg steering weel, speed) as input and renders the next screen.
        It allows users to control different aspects of the scene, such as weather and objects.
    }
    \label{fig:overview}
        \vspace{-5mm}
\end{figure}

Data-driven simulation has recently gained attention. LidarSim~\cite{manivasagam2020lidarsim} used a catalog of annotated 3D scenes to sample layouts into which reconstructed objects obtained from a large number of recorded drives are placed, in the quest to achieve diversity for training and testing a LIDAR-based perception system. 
~\cite{kar2019meta,devaranjan2020meta,ruiz2018learning}, on the other hand, learn to synthesize road-scene 3D layouts directly from images without supervision. These works do not model the dynamics of the environment and object behaviors.

As a more daring alternative, recent works attempted to create neural simulators~\cite{kim2020learning,ha2018recurrent} that learn to simulate the environment in response to the agent's actions directly in pixel-space by digesting large amounts of video data along with actions.
This line of work provides a scaleable way to simulation, as we do not rely on any human-provided annotations, except for the agent's actions which are cheap to obtain from odometry sensors. It is also a more challenging way, since the complexity of the world and the dynamic agents acting inside it, needs to be learned in a high-resolution camera view. In this paper, we follow this route.


We introduce {\Name}, a neural simulator that learns from sequences of video footage and associated actions 
taken by an ego-agent in an environment.
{\Name} leverages Variational-Auto Encoder~\cite{kingma2014autoencoding} and Generative Adversarial Networks~\cite{gan} to learn a latent space for images on which a dynamics engine learns the transitions within the latent space.
The key aspects of {\Name} are its disentangled latent space and high-resolution and high-fidelity frame synthesis conditioned on the agent's actions. The disentanglement property of {\Name} gives users additional control over the environment, such as changing the weather and locations of non-player objects. Furthermore, since {\Name} is an end-to-end differentiable simulator, we are able to re-create the scenarios observed from real video footage allowing the agent to drive again through the recorded scene but taking different actions. This property makes {\Name} the first neural driving simulator of its kind.
By learning on 160 hours of real driving data, we showcase {\Name} to learn high-fidelity simulation, surpassing all existing neural simulators by a significant margin, and allowing for the control over the environment not possible previously.

%

\section{Related Work}

\subsection{Video Generation and Prediction}
As in image generation, the standard architectures for video generation are VAEs \cite{DBLP:journals/corr/abs-1802-07687,DBLP:journals/corr/abs-1806-04166}, auto-regressive models \cite{DBLP:journals/corr/RanzatoSBMCC14,DBLP:journals/corr/SrivastavaMS15,DBLP:journals/corr/KalchbrennerOSD16,DBLP:journals/corr/abs-1906-02634}, flow-based models~\cite{DBLP:journals/corr/abs-1903-01434}, and GANs \cite{mathieu2016deep,vondrick2016generating,saito2017temporal,Saito2018TGANv2ET,dvdgan,mocogan}.
For a generator to sample videos, it must be able to generate realistic looking frames as well as realistic transitions between frames.
Video prediction models~\cite{oh2015action,lotter2016deep,finn2016unsupervised,minderer2019unsupervised,DBLP:journals/corr/abs-1710-11252,lee2018stochastic,crevnet20} learn to produce future frames given a reference frame, and they share many similarities to video generation models.
Similar architectures can be applied to the task of conditional video generation in which information such as semantic segmentation is given as input to the model \cite{vid2vid, mallya2020worldconsistent}. In this work, we use a VAE-GAN \cite{DBLP:journals/corr/LarsenSW15} based on StyleGAN \cite{karras2019style} to learn a latent space of natural images, then train a dynamics model within the space.


\subsection{Data-driven Simulation and Model-based RL}
The goal of data-driven simulation is to learn simulators given observations from the environment to be simulated.
Meta-Sim~\cite{kar2019meta,devaranjan2020meta} learns to produce scene parameters in a synthetic scene.
LiDARSim~\cite{manivasagam2020lidarsim} leverages deep learning and physics engine to produce LiDAR point clouds.
In this work, we focus on data-driven simulators that produce future frames given controls.
World Model \cite{ha2018recurrent} use a VAE~\cite{kingma2014autoencoding} and LSTM~\cite{hochreiter1997long} to model transition dynamics and rendering functionality.
In GameGAN \cite{kim2020learning}, a GAN and a memory module are used to mimic the engine behind games such as Pacman and VizDoom.
Model-based RL~\cite{sutton1990integrated,deisenroth2011pilco,hafner2019learning,kaiser2019model,ha2018recurrent} also aims at learning a dynamics model of some environment which agents can utilize to plan their actions.
While prior work has applied neural simulation to simple environments~\cite{bellemare2013arcade,todorov2012mujoco} in which a ground-truth simulator is already known, we also apply our model to real-world driving data and focus on improving the quality of simulations.
Furthermore, we show how users can interatively edit scenes to create diverse simulation environments.


%

\section{Methodology}

\begin{figure}[t!]
    \vspace{-1mm}
    \begin{center}
        \includegraphics[width=\linewidth]{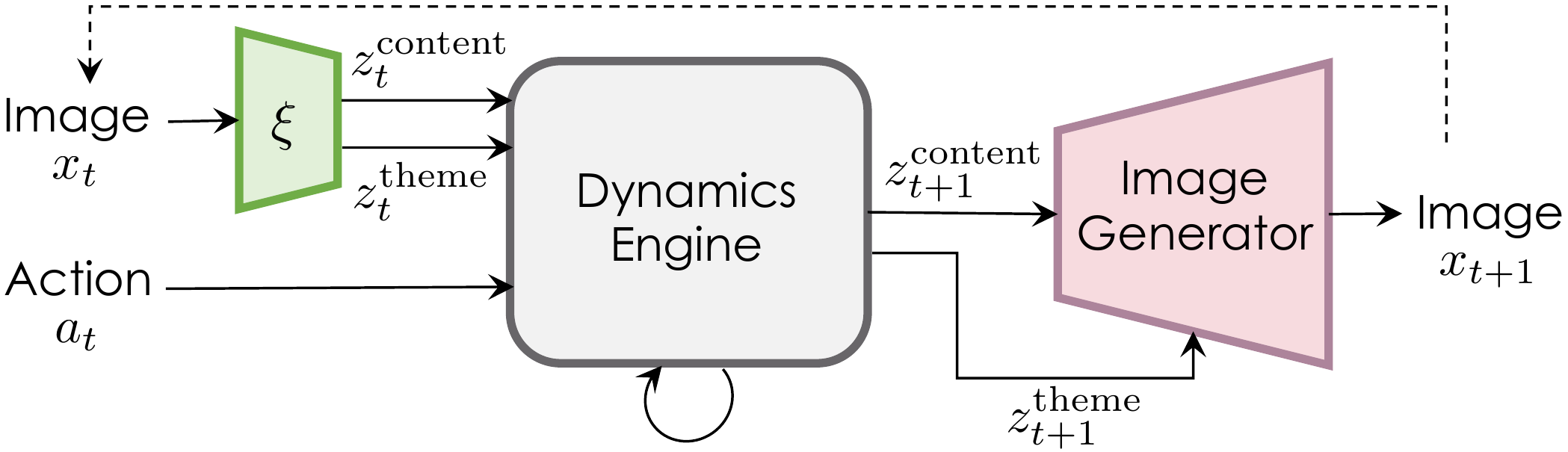}
    \end{center}
    \vspace{-4.5mm}
    \caption[]{
        \small {\Name} takes an image $x_t$ and action $a_t$ as input at time $t$.
        With  encoder $\xi$, $x_t$ is encoded into disentangled latent codes $z_t^{\text{theme}}$ and $z_t^{\text{content}}$.
        Dynamics Engine learns the transition function for the latent codes given $a_t$.
        Image Generator produces $x_{t+1}$, which is fed to the next time step, autoregressively.
    }
    \label{fig:system_overview}
    \vspace{-1.5mm}
\end{figure}

Our objective is to learn a high-quality controllable neural simulator by watching sequences of video frames and their associated actions.
We aim to achieve controllability in two aspects:
1) We assume there is an egocentric agent that can be controlled by a given action.
2) We want to control different aspects of the current scene, for example, by modifying an object or changing the background color.

\begin{figure}[t!]
    \vspace{-1mm}
    \begin{center}
        \includegraphics[width=0.96\linewidth]{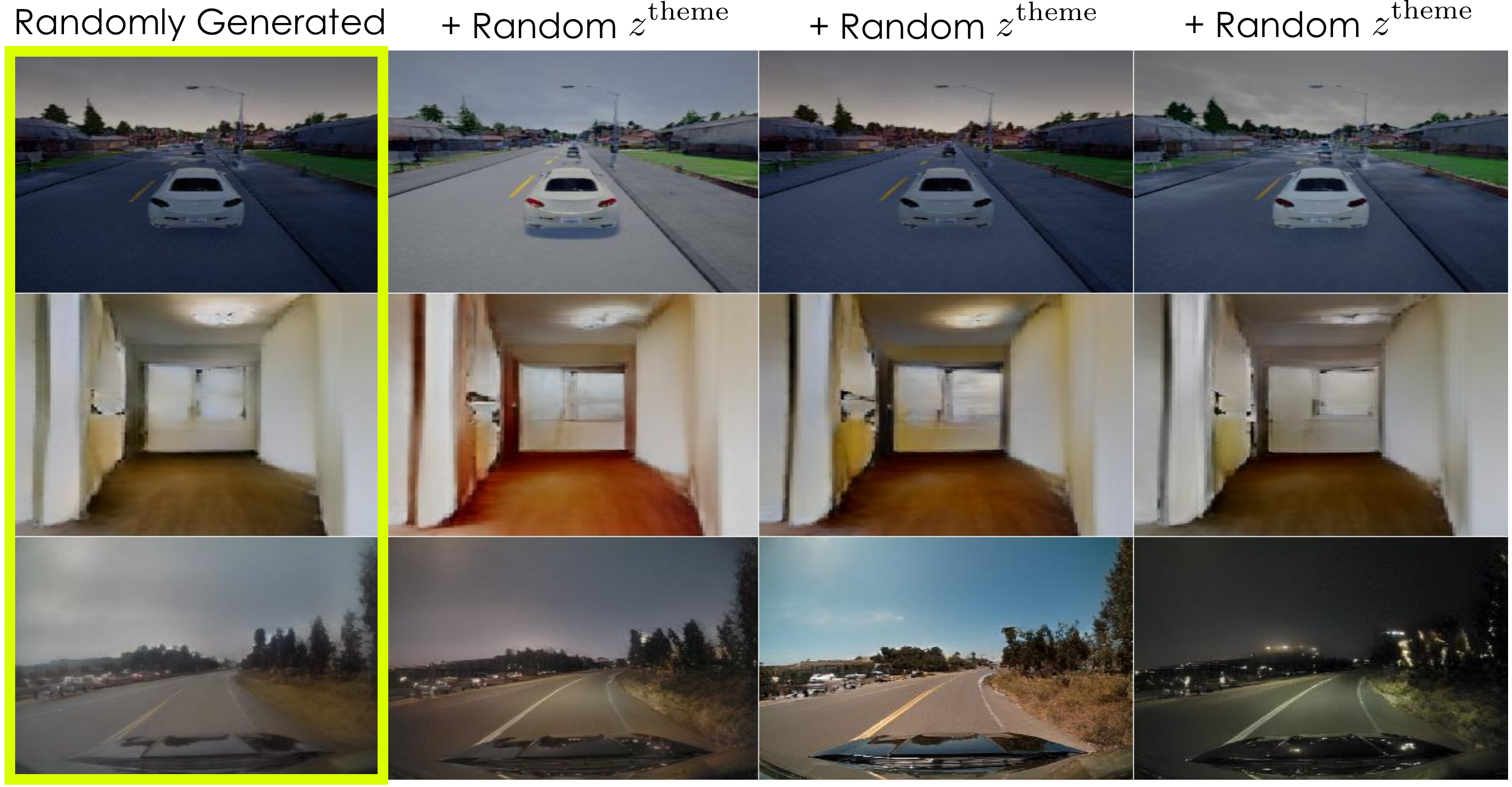}
    \end{center}
    \vspace{-6mm}
    \caption[]{
    \small Left column shows randomly generated images from different environments.
    By sampling $z^{\text{theme}}$, we can change theme information such as weather while keeping the content consistent.
    }
    \label{fig:change_theme}
    \vspace{-3.5mm}
\end{figure}

Let us denote the video frame at time $t$ as $x_t$ and the continuous action as $a_t$.
We learn to produce the next frame $x_{t+1}$ given the previous frames $x_{1:t}$ and actions $a_{1:t}$.
Fig~\ref{fig:system_overview} provides an overview of our model.
Image encoder $\xi$ produces the disentangled latent codes $z^{\text{theme}}$ and $z^{\text{content}}$ for $x$ in an unsupervised manner.
We define \textit{\text{theme}} as information that does not depend on pixel locations such as the background color or weather of the scene, and \textit{\text{content}} as spatial content (Fig~\ref{fig:change_theme}).
Dynamics Engine, a recurrent neural network, learns to produce the next latent codes $z_{t+1}^{\text{theme}}$, $z_{t+1}^{\text{content}}$ given $z_t^{\text{theme}}$, $z_t^{\text{content}}$, and $a_t$.
$z_{t+1}^{\text{theme}}$ and $z_{t+1}^{\text{content}}$ go through an image decoder that generates the output image.

Generating high-quality temporally-consistent image sequences is a challenging problem~\cite{DBLP:journals/corr/abs-1903-01434,mathieu2016deep,dvdgan,mocogan,vid2vid,mallya2020worldconsistent}.
Rather than generating a sequence of frames directly, we split the learning process into two steps, motivated by World Model~\cite{ha2018recurrent}.
Sec~\ref{sec:pretrain} introduces our encoder-decoder architecture that is pre-trained to produce the latent space for images.
We propose a novel architecture that disentangles themes and content while achieving high-quality generation by leveraging a Variational Auto-Encoder (VAE) 
and Generative Adversarial Networks (GAN). 
Sec~\ref{sec:dynamics_engine} describes the Dynamics Engine that learns the latent space dynamics.
We also show how the Dynamics Engine further disentangles action-dependent and action-independent content.

\subsection{Pre-trained Latent Space}
\label{sec:pretrain}
We build our image decoder on top of the popular StyleGAN~\cite{karras2019style,karras2020analyzing}, but make several modifications that allow for theme-content disentanglement.
Since extracting the GAN's latent code that corresponds to an input image is not trivial, we introduce an encoder $\xi$ that maps an image $x$  into its latent code $z$. 
We utilize the VAE formulation, particularly the $\beta$-VAE~\cite{higgins2016beta} to control the KL term better.
Therefore, on top of the adversarial losses from StyleGAN, we add the following loss at each step of generator training:
\begin{equation*}
\label{eqn:vae}
L_{VAE} = E_{z\sim q(z|x)}[log(p(x|z))] + \beta KL(q(z|x) || p(z))
\end{equation*}
where $p(z)$ is the standard normal prior distribution, $q(z|x)$ is the approximate posterior from the encoder $\xi$,
and $KL$ is the Kullback-Leibler divergence.
For the reconstruction  term, we reduce the perceptual distance~\cite{zhang2018unreasonable} between the input and output images rather than the pixel-wise distance.

\begin{figure}[t!]
    \vspace{-1mm}
    \begin{center}
        \includegraphics[width=\linewidth]{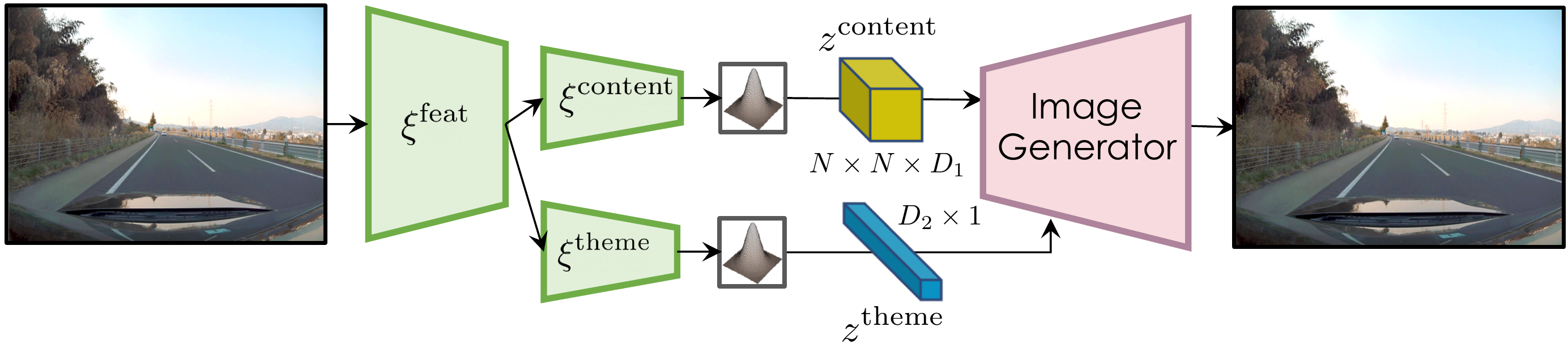}
    \end{center}
    \vspace{-5.5mm}
    \caption[]{
        \small Pretraining stage learns the encoder and decoder for images.
        The encoder $\xi$ produces $z^{\text{content}}$ and $z^{\text{theme}}$ which comprise the disentangled latent space that the dynamics engine trains on.
        The gaussian blocks represent reparameterization steps~\cite{kingma2014autoencoding}.
    }
    \label{fig:model_pretraining}
    \vspace{-2.5mm}
\end{figure}

This form of combining VAE and GAN has been explored before \cite{DBLP:journals/corr/LarsenSW15}.
To achieve our goal of controllable simulation, we introduce several novel modifications to the encoder and decoder.
Firstly, we disentangle the \textit{\text{theme}} and \textit{\text{content}} of the input image.
Our encoder $\xi$ is composed of a feature extractor $\xi^{\text{feat}}$ and two encoding heads $\xi^{\text{content}}$ and $\xi^{\text{theme}}$ (Figure ~\ref{fig:model_pretraining}).
$\xi^{\text{feat}}$ takes an image $x$ as input and consists of several convolution layers whose output is passed to the two heads.
$\xi^{\text{content}}$ produces $z^{\text{content}} \in \mathbb{R}^{N \times N \times D_1}$ which has $N \times N$ spatial dimension.
On the other hand, $\xi^{\text{theme}}$ produces $z^{\text{theme}} \in \mathbb{R}^{D_2}$, a single vector, which controls the theme of the output image.
Let us denote $z = \{z^{\text{content}}, z^{\text{theme}}\}$.
Note that $z^{\text{content}}$ and $z^{\text{theme}}$ are matched to be from the standard normal prior by the reparametrization and training of VAE.
We feed $z$ into the StyleGAN decoder.
StyleGAN controls the appearance of generated images with adaptive instance normalization ($AdaIN$)~\cite{dumoulin2016learned,huang2017arbitrary,ghiasi2017exploring} layers after each convolution layer of its generator.
$AdaIN$ applies the same scaling and bias to each spatial location of a normalized feature map:
\begin{equation}
\vspace{-2mm}
AdaIN(\mathbf{m}, \alpha, \gamma) = \mathcal{A}(\mathbf{m}, \alpha, \gamma) = \alpha \frac{\mathbf{m} - \mu(\mathbf{m})}{\sigma(\mathbf{m})} + \gamma
\end{equation}
where $\mathbf{m} \in \mathbb{R}^{N \times N \times 1}$ is a feature map with $N\times N$ spatial dimension and $\alpha, \gamma$ are scalars for scaling and bias.
Thus, $AdaIN$ layers are perfect candidates for inserting \textit{\text{theme}} information.
We pass $z^{\text{theme}}$ through an $MLP$ to get the scaling and bias values for each $AdaIN$ layer.
Now, because of the shape of $z^{\text{content}}$, it naturally encodes the content information from the corresponding $N\times N$ grid locations.
Rather than having a constant block as the input to the first layer as in StyleGAN, we pass $z^{\text{content}}$ as the input.
Furthermore, we can sample a new vector $v \in \mathbb{R}^{1 \times 1 \times D_1}$ from the normal prior distribution to swap out the content of some grid location.
Preliminary experiments showed that encoding information only using the plain StyleGAN decoder is not adequate for capturing the details of scenes with multiple objects because the generator must recover spatial information from the inputs to $AdaIN$ layers, which apply the same scaling and bias to all spatial locations.
We use the multi-scale multi-patch discriminator architecture~\cite{vid2vid,isola2017image,shaham2019singan}, which results in higher quality images for complex scenes.
We use the same adversarial losses $L_{GAN}$ from StyleGAN, and the final loss function is $L_{pretrain} = L_{VAE} + L_{GAN}$.

We observe that balancing the KL loss with suitable $\beta$ in $L_{VAE}$ is essential.
Smaller $\beta$ gives better reconstruction quality, but the learned latent space could be far away from the prior,
in which case the dynamics model (Sec.\ref{sec:dynamics_engine}) had a harder time learning the dynamics.
This causes $z$ to be overfit to $x$, and it becomes more challenging to learn the transitions between frames in the overfitted latent space.

\vspace{0mm}
\subsection{Dynamics Engine}
\label{sec:dynamics_engine}
With the pre-trained encoder and decoder, the Dynamics Engine learns the transition between latent codes from one time step to the next given an action $a_t$.
We fix the parameters of the encoder and decoder, and only learn the parameters of the engine.
This allows us to pre-extract latent codes for a dataset before training.
The training process becomes faster and significantly easier than directly working with images, as latent codes typically have dimensionality much smaller than the input.
In addition, we further disentangle \textit{\text{content}} information from $z^{\text{content}}$ into \textit{action-dependent} and \textit{action-independent} features without supervision.

In a 3D environment, the view-point 
shifts as the ego agent moves.
This shifting naturally happens spatially, so we employ a convolutional LSTM module (Figure~\ref{fig:model_dynamics_engine}) to learn the spatial transition between each time step:
\begin{equation}
    v_t = \mathcal{F}(\mathcal{H}(h^{\text{conv}}_{t-1}, a_t, z_t^{\text{content}}, z_t^{\text{theme}}))
\end{equation}
\begin{equation}
    i_t, f_t, o_t = \sigma(v_t^i),\sigma(v_t^f), \sigma(v_t^o)
\end{equation}
\begin{equation}
    c^{\text{conv}}_{t} = f_t \odot c^{\text{conv}}_{t-1} + i_t \odot \tanh(v_t^g)
\end{equation}
\begin{equation}
    h^{\text{conv}}_{t} = o_t \odot \tanh(c^{\text{conv}}_{t})
\end{equation}
where $h^{\text{conv}}_{t}, c^{\text{conv}}_{t}$ are the hidden and cell state of the convLSTM module, and $i_t, f_t, o_t$ are the input, forget, output gates, respectively.
$\mathcal{H}$ replicates $a_t$ and $z_t^{\text{theme}}$ spatially to match the $N\times N$ spatial dimension of $z^{\text{content}}_t$.
It fuses all inputs by concatenating and running through a $1\times 1$ convolution layer.
$\mathcal{F}$ is composed of two $3\times 3$ convolution layers.
$v_t$ is split into intermediate variables $v_t^i, v_t^f, v_t^o, v_t^g$.
All state and intermediate variables have the same size $\mathbb{R}^{N\times N\times D_{\text{conv}}}$.
The hidden state $h^{\text{conv}}_{t}$ goes through two separate convolution layers to produce $z_{t+1}^{\text{theme}}$ and $z_{t+1}^{a_{\text{dep}}}$.
The action dependent feature $z_{t+1}^{a_{\text{dep}}}$ is used to produce $z_{t+1}^{\text{content}}$, along with $z_{t+1}^{a_{\text{indep}}}$. 

\begin{figure}[t!]
    \begin{center}
        \includegraphics[width=0.92\linewidth]{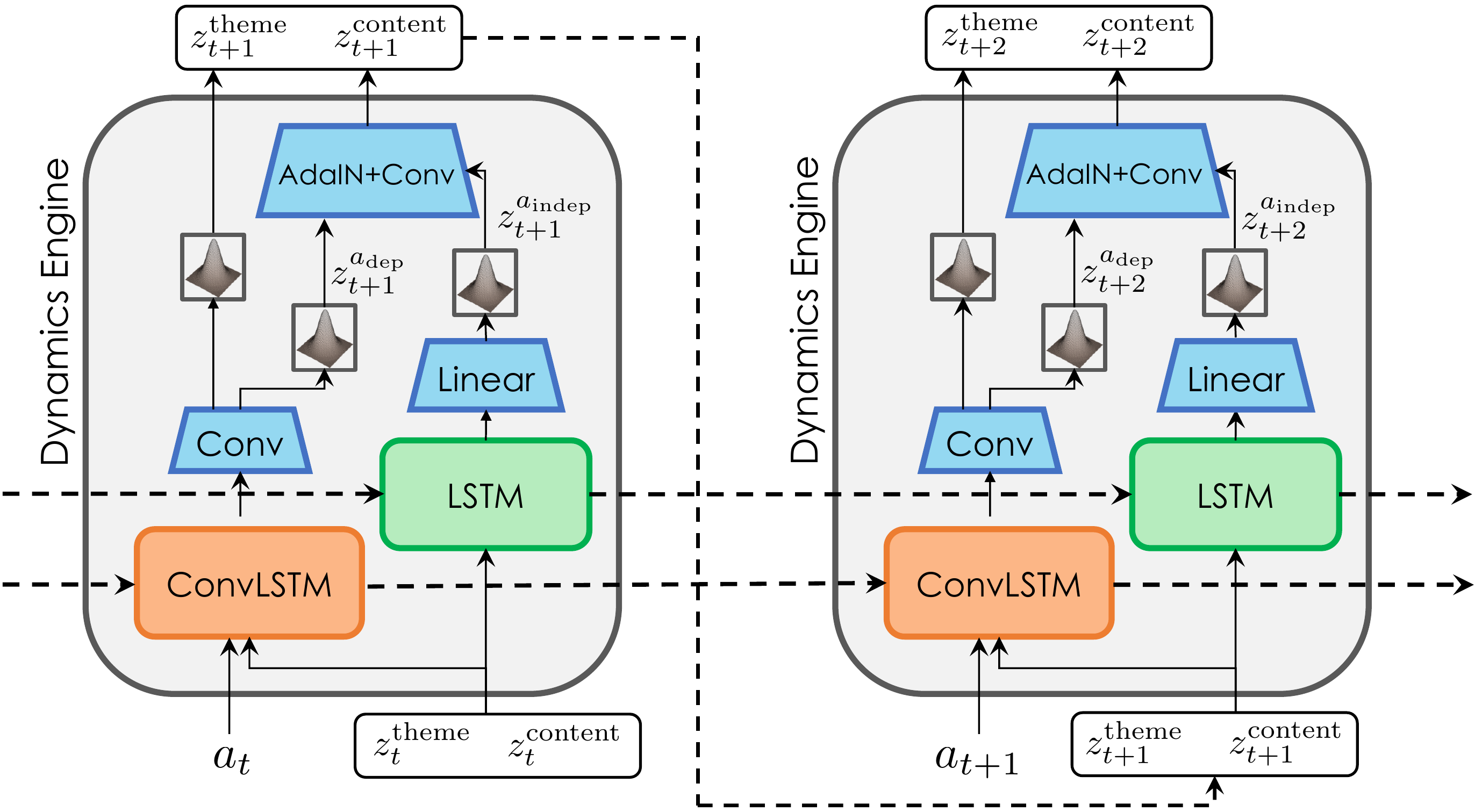}
    \end{center}
    \vspace{-5.5mm}
    \caption[]{
        \small Dynamics Engine produces the next latent codes, given an action and previous latent codes.
        It disentangles content information into action-dependent and action-independent features with its two separate LSTMs.
        Dashed lines correspond to temporal connections.
         Gaussian blocks indicate reparameterization steps. 
    }
    \label{fig:model_dynamics_engine}
    \vspace{-1.5mm}
\end{figure}

We also add a plain LSTM~\cite{hochreiter1997long} module that only takes $z_t$ as input.
Therefore, this module is responsible for information that does not depend on the action $a_t$.
The input $z_t$ is flattened into a vector, and all variables inside this module have size $\mathbb{R}^{D_{\text{linear}}}$.
The hidden state goes through a linear layer that outputs $z_{t+1}^{a_{\text{indep}}}$.
Finally, $z_{t+1}^{a_{\text{dep}}}$ and $z_{t+1}^{a_{\text{indep}}}$ are used as inputs to two $AdaIN$ + Conv blocks.
\begin{equation}
    \bm{\alpha}, \bm{\beta} = MLP(z_{t+1}^{a_{\text{indep}}})
\end{equation}
\begin{equation}
    z_{t+1}^{\text{content}} = \mathcal{C}(\mathcal{A}(\mathcal{C}(\mathcal{A}(z_{t+1}^{a_{\text{dep}}}, \bm{\alpha}, \bm{\beta})),\bm{\alpha}, \bm{\beta}))
\end{equation}
where we denote convolution and $AdaIN$ layers as $\mathcal{C}$ and $\mathcal{A}$, respectively.
An $MLP$ is used to produce $\bm{\alpha}$ and $\bm{\beta}$.
We reparameterize $z^{a_{\text{dep}}}$,$z^{a_{\text{indep}}}$, $z^{\text{theme}}$ into the standard normal distribution $N(0,I)$ which allows sampling at test time:
\begin{equation}
\label{eq:reparam}
    z = \mu + \epsilon \sigma, \quad \epsilon \sim N(0, I)
\end{equation}
where $\mu$ and $\sigma$ are the intermediate variables for the mean and standard deviation for each reparameterization step.

Intuitively, $z^{a_{\text{indep}}}$ is used as $style$ for the spatial tensor $z^{a_{\text{dep}}}$ through $AdaIN$ layers.
$z^{a_{\text{indep}}}$ does not get action information, so it alone cannot learn to generate plausible next frames.
This architecture thus allows disentangling action-dependent features such as the layout of a scene from action-independent features such as object types.
Note that the engine could ignore $z^{a_{\text{indep}}}$ and only use $z^{a_{\text{dep}}}$ to learn dynamics.
If we keep the model size small and use a high $KL$ penalty on the reparameterized variables, it will utilize full model capacity and make use of $z^{a_{\text{indep}}}$.
We can also enforce disentanglement between $z^{a_{\text{indep}}}$ and $z^{a_{\text{dep}}}$ using an adversarial loss~\cite{denton2017unsupervised}.
In practice, we found that our model was able to disentangle information well without such a loss.

\textbf{Training:}
We extend the training procedure of GameGAN~\cite{kim2020learning} in latent space to train our model with adversarial and VAE losses.
Our adversarial losses $L_{adv}$ come from two networks: 1) single latent discriminator, and 2) temporal action-conditioned discriminator.
We first flatten $z_t$ into a vector with size $\mathbb{R}^{N^2D_1+D_2}$.
The single latent discriminator is an $MLP$ that tries to discriminate produced $z_t$ from the real latent codes.
The temporal action-conditioned discriminator is implemented as a temporal convolution network such that we apply filters in the temporal dimension~\cite{kim2014convolutional} where the actions $a_t$ are fused to the temporal dimension.
We also sample negative actions $\bar{a}_t$, and the job of the discriminator is to figure out if the given sequence of latent codes is realistic and faithful to the given action sequences.
We use the temporal discriminator features to reconstruct the input action sequence and reduce the action reconstruction loss $L_{action}$ to help the dynamics engine to be faithful to the given actions.
Finally, we add latent code reconstruction loss $L_{latent}$ so that the generated $z_t$ matches the ground truth latent codes, and reduce the $KL$ penalty $L_{KL}$ for $z_t^{a_{\text{dep}}}$,$z_t^{a_{\text{indep}}}$, $z_t^{\text{theme}}$.
The final loss function is $L_{DE} = L_{adv} + L_{latent} + L_{action} + L_{KL}$.
Our model is trained with 32 time-steps with a warm-up phase similar to GameGAN. Further details are provided in the Appendix.

\subsection{Differentiable Simulation}
\label{sec:diff_sim}
One compelling aspect of {\Name} is that it can create an editable simulation environment from a real video.
As {\Name} is fully differentiable, it allows for recovering the scene and scenario by discovering the underlying factors of variations that comprise a video, while also recovering the actions that the agent took, if these are not provided.  We refer to this as \textit{differentiable simulation}.
Once these parameters are discovered, the agent can use {\Name} to re-simulate the scene and take different actions. {\Name} further allows sampling and modification of various components of a scene, thus testing the agent in the same scenario under different weather conditions or objects. 

\begin{figure}[t!]
\vspace{-1mm}
    \begin{center}
        \includegraphics[width=\linewidth]{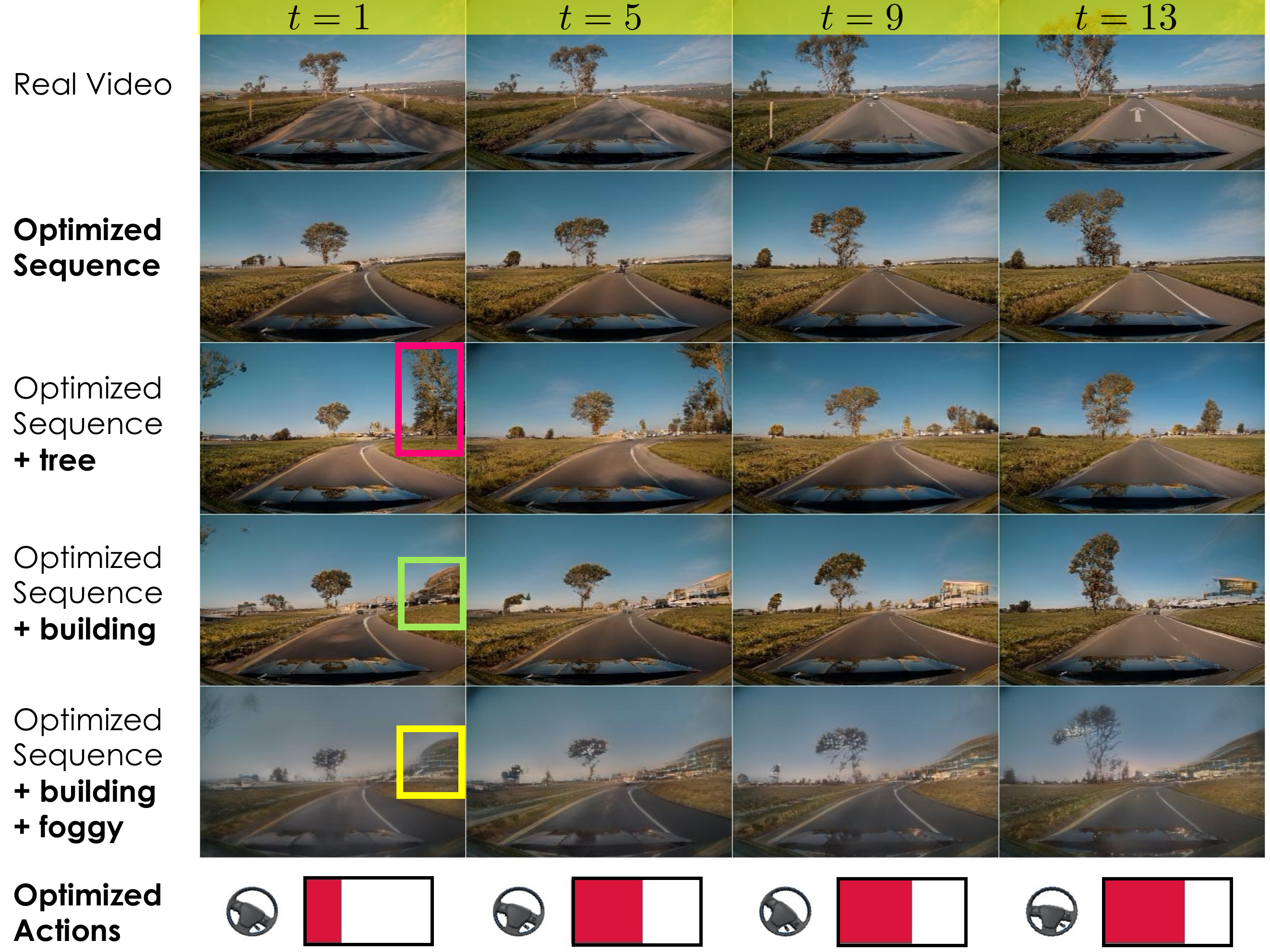}
    \end{center}
    \vspace{-4.5mm}
    \caption[]{
        \textbf{Differentiable simulation}:
        \small We can first optimize for the underlying sequence of inputs that can reproduce a real video.
        With its controllability, we can replay the same scenario with modified content or scene condition.
    }
    \label{fig:diffSim_pilotnet}
    \vspace{-2.5mm}
\end{figure}

First, note that reparametrization steps (Eq.~\ref{eq:reparam}) involve a stochastic variable $\epsilon$ which gives stochasticity in a simulation to produce diverse future scenarios.
Given a sequence of frames from a real video $x_0, ..., x_T$, our model can be used to find the underlying $a_0,...,a_{T-1}, \epsilon_0,...\epsilon_{T-1}$:
\begin{equation}
\label{eq:diffSim}
    \underset{a_{0..T-1}, \epsilon_{0..T-1}}{\text{minimize}} \sum_{t=1}^{T} ||z_t - \hat{z}_t || + \lambda_1||a_t - a_{t-1}|| + \lambda_2||\epsilon_t||
\end{equation}
where $z_t$ is the output of our model, $\hat{z}_t$ is the encoding of $x_t$ with the encoder, and $\lambda_1,\lambda_2$ are hyperparameters for regularizers.
We add action regularization assuming the action space is continuous and $a_t$ does not differ significantly from $a_{t-1}$.
To prevent the model from utilizing $\epsilon_t$ to explain all differences between frames, we also add the $\epsilon$ regularizer.

\begin{figure}[b!]
    \vspace{-2mm}
    %
    \begin{minipage}{0.61\linewidth}
    \begin{center}
        \includegraphics[width=\linewidth]{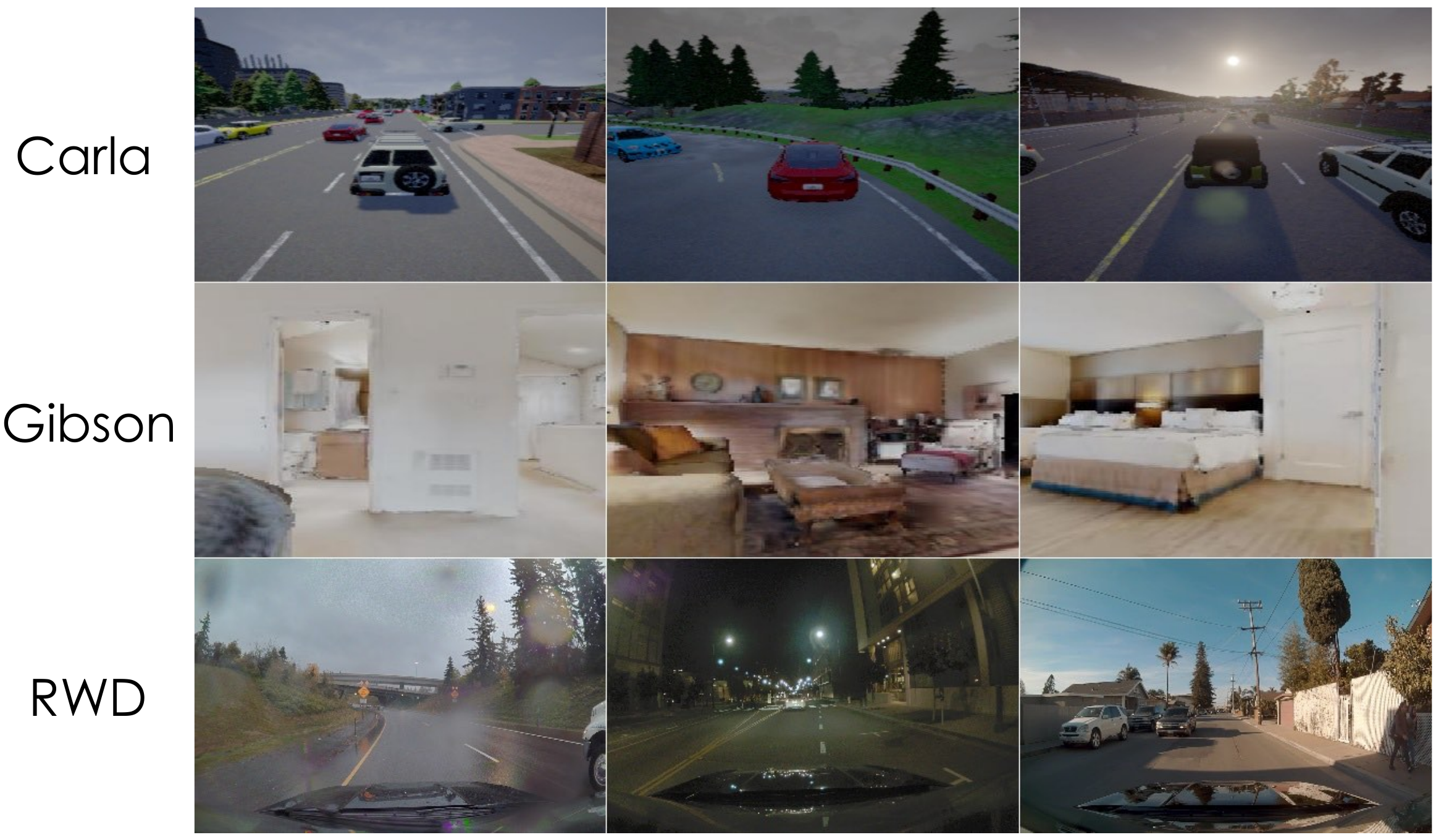}
    \end{center}
    \end{minipage}
    \hspace{0.5mm}
    \begin{minipage}{0.35\linewidth}
    \vspace{2.5mm}
    \caption[]{
    \small Image samples from three datasets studied in this work, for simulated and real-world driving, and indoor navigation.
    }
    \label{fig:datasets}
    \end{minipage}
    \vspace{-2mm}
\end{figure}

\begin{figure*}[t!]
\vspace{-1mm}
    \begin{center}
        \includegraphics[width=0.87\textwidth]{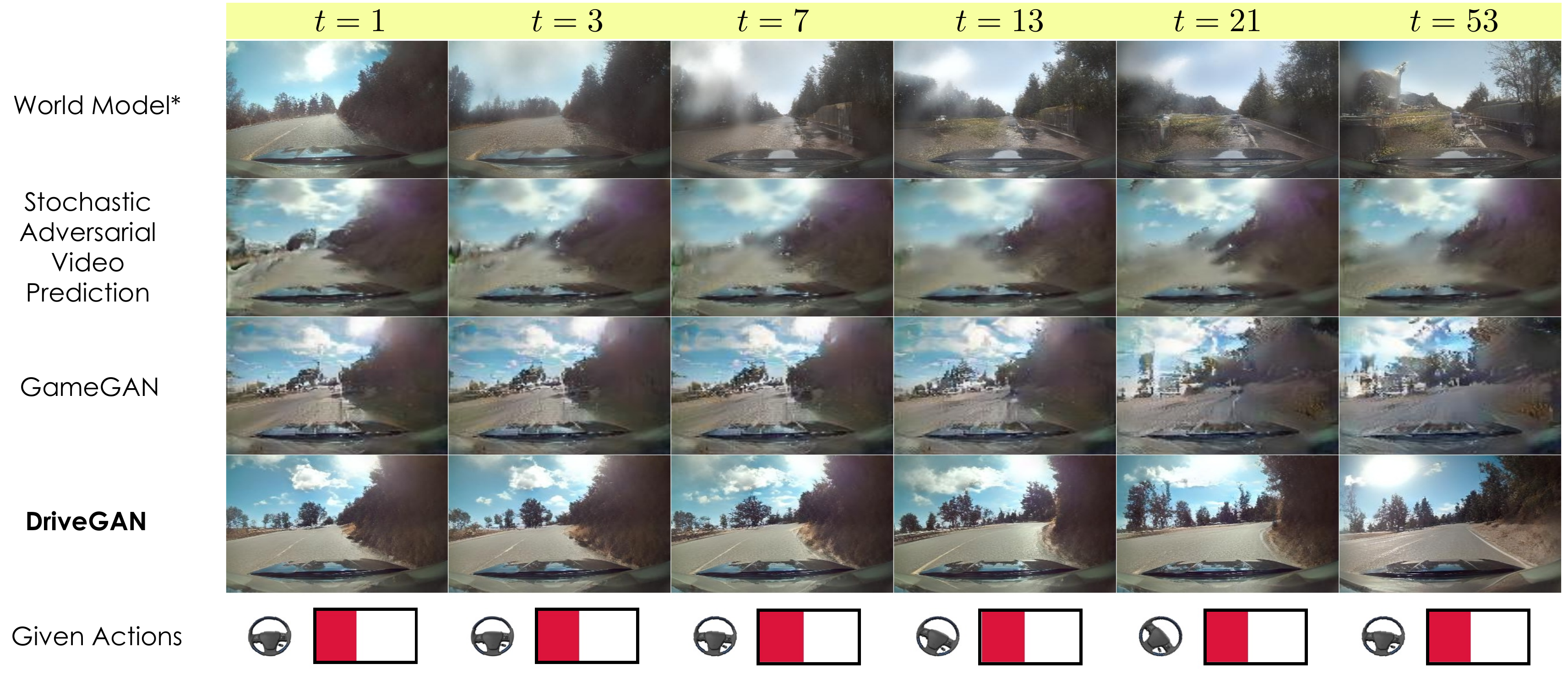}
    \end{center}
    \vspace{-5.5mm}
    \caption[]{
        \small Comparison of baseline models.
        All models are given the same initial screen and sequence of actions.
        Our model can produce a high-quality temporally consistent simulation that conforms to the action sequence.
    }
    \label{fig:baseline_comparison}
     \vspace{-2mm}
\end{figure*}

\section{Experiments}

%

We perform thorough quantitative (Sec~\ref{sec:quant_results}) and qualitative (Sec~\ref{sec:qual_results}) experiments on the following datasets.

\textbf{Carla}~\cite{Dosovitskiy17}
simulator is an open-source simulator for autonomous driving research.
We use five towns in Carla to generate the dataset.
The ego-agent and other vehicles are randomly placed and use random policy to drive in the environment.
Each sequence has a randomly sampled weather condition and consists of 80 frames sampled at 4Hz.
48K sequences are extracted, and 43K are used for training.

\textbf{Gibson}
environment~\cite{xiazamirhe2018gibsonenv} 
virtualizes real-world indoor buildings and has an integrated physics engine with which virtual agents can be controlled.
We first train a reinforcement learning agent that can navigate towards a given destination coordinate.
In each sequence, we randomly place the agent in a building and sample a destination.
85K sequences each with 30 frames are extracted from 100 indoor environments, and 76K sequences are used for training.

\textbf{Real World Driving} (RWD)
data consists of real-world recordings of human driving on multiple different highways and cities.
It was collected in a variety of different weather and times.
RWD is composed of 128K sequences each with 36 frames extracted at 8Hz.
It corresponds to $\sim$ 160 hours of driving, and we use 125K sequences for training.

Figure~\ref{fig:datasets} illustrates scenes from the datasets.
Each sequence consists of the extracted frames (256$\times$256) and the actions the ego agent takes at each time step.
The 2-dim actions consist of the agent's speed and angular velocity.

\begin{figure}[!h]
    \vspace{-3mm}
    \begin{center}
        \includegraphics[width=0.76\linewidth]{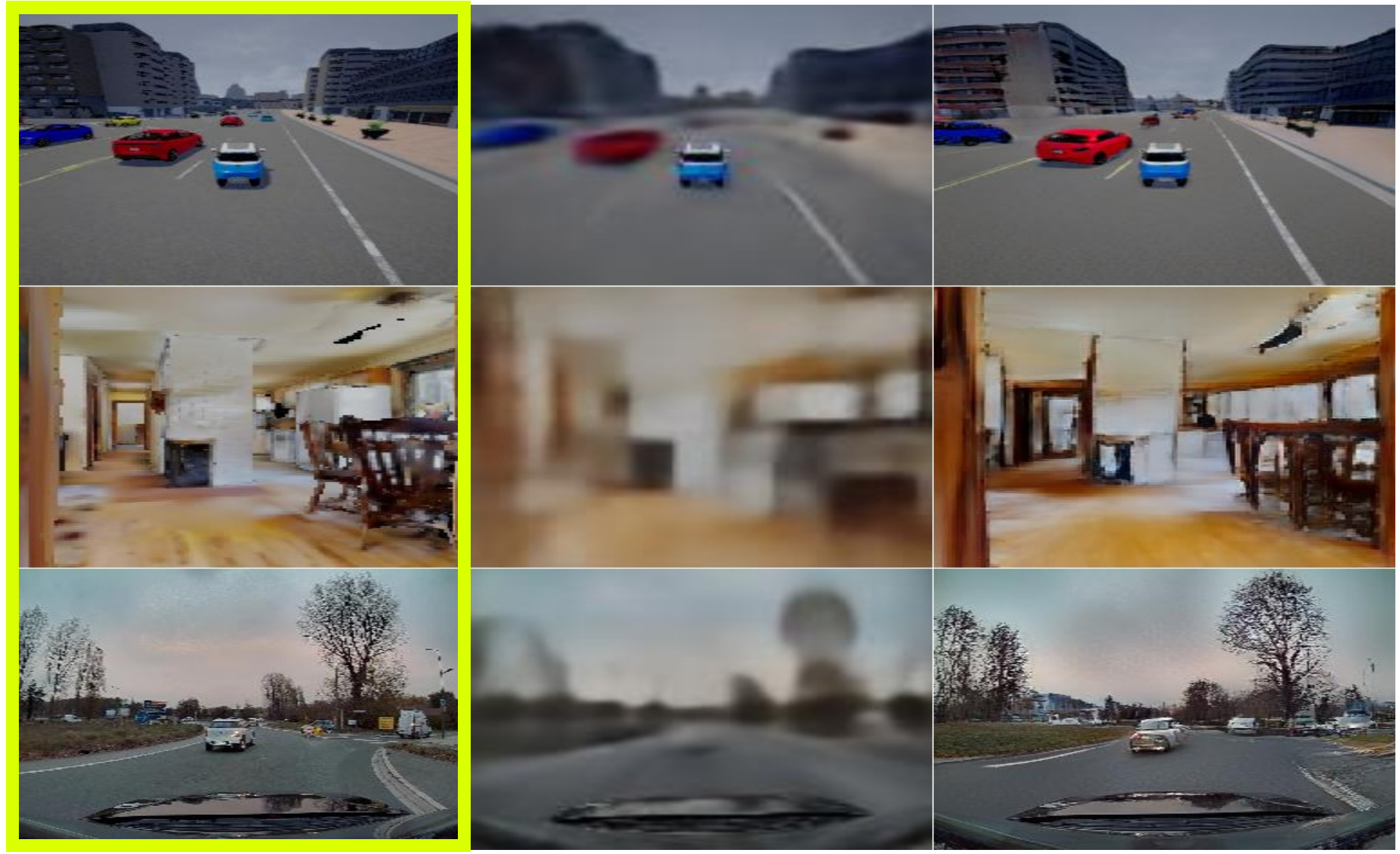}
    \end{center}
    \vspace{-6mm}
    \caption[]{
    \footnotesize
    \textbf{Left:} original images, \textbf{Middle:} reconstructed images from VAE, \textbf{Right:} reconstructed images from our encoder-decoder model.
    }
    \label{fig:vae_comparison}
    \vspace{-5mm}
\end{figure}

\subsection{Quantitative Results}
\label{sec:quant_results}
The quality of simulators needs to be evaluated in two aspects.
The generated videos from simulators have to look realistic, and their distribution should match the distribution of the real videos.
They also need to be faithful to the action sequences used to produce them.
This is essential to be useful for downstream tasks, such as training a robot.
Therefore, we use two automatic metrics to measure the performance of models.
The experiments are carried out by using the first frames and action sequences of the test set.
The remaining frames are generated autoregressively.

We compare with four baseline models:
Action-RNN~\cite{chiappa2017recurrent} is a simple action-conditioned RNN model trained with reconstruction loss on the pixel space,
Stochastic Adversarial Video Prediction (SAVP)~\cite{lee2018stochastic} and GameGAN~\cite{kim2020learning} are trained with adversarial loss along with reconstruction loss on the pixel space,
World Model~\cite{ha2018recurrent} trains a vision model based on VAE and an RNN based on mixture density networks (MDN-RNN).
World Model is similar to our model as they first extract latent codes and learn MDN-RNN on top of the learned latent space.
However, their VAE is not powerful enough to model the complexities of the datasets studied in this work.
Fig~\ref{fig:vae_comparison} shows how a simple VAE cannot reconstruct the inputs; thus, the plain World Model cannot produce realistic video sequences by default.
Therefore, we include a variant, denoted as World Model*, that uses our proposed latent space to train the MDN-RNN component.

\begin{figure}[h!]
    \vspace{0mm}
    \begin{center}
        \includegraphics[width=0.88\linewidth]{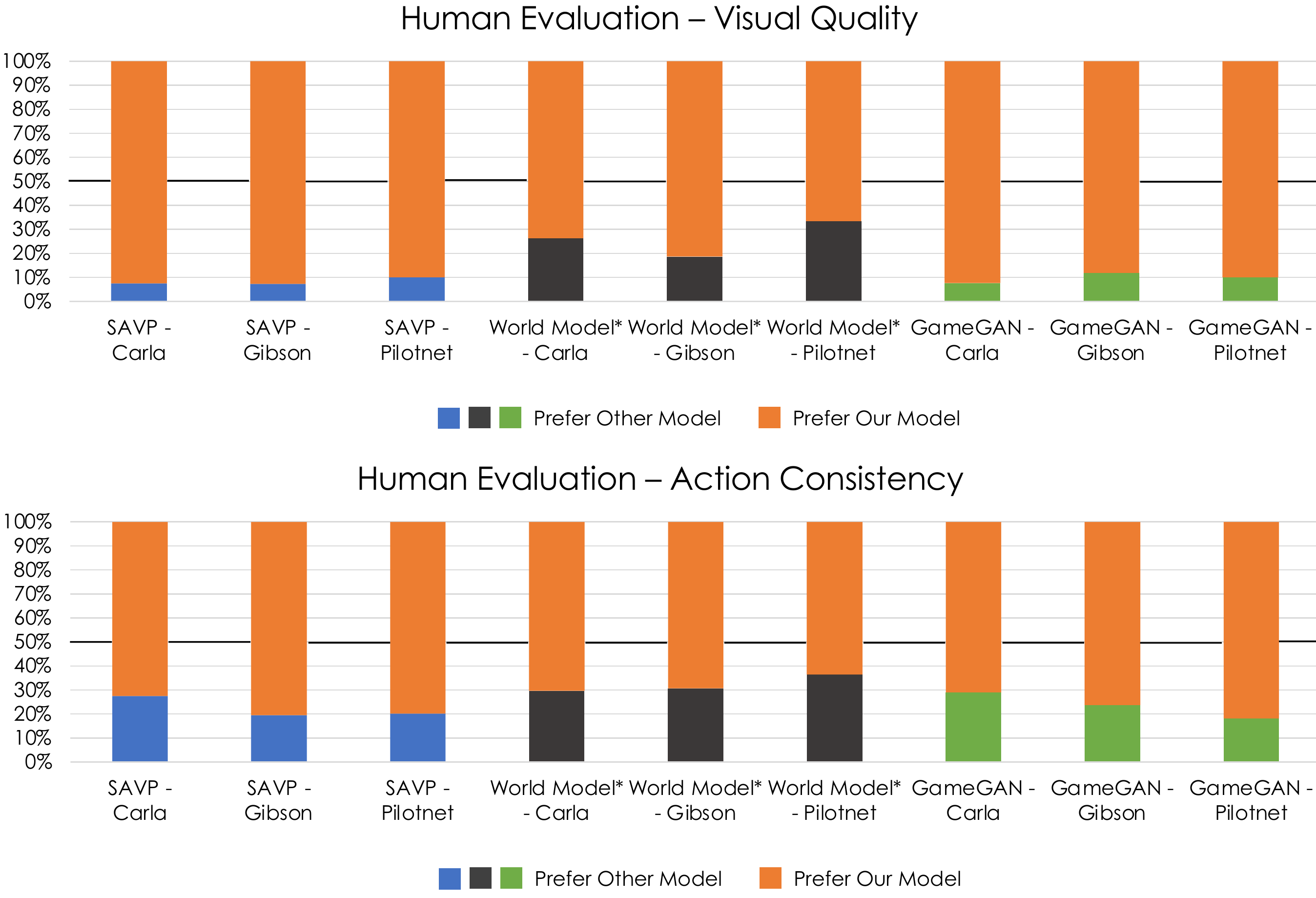}
    \end{center}
    \vspace{-5mm}
    \caption[]{
  \small  \textbf{Human evaluation:} Our model outperforms baseline models on both criteria.
    }
    \label{fig:human_evaluation_all}
    \vspace{-5mm}
\end{figure}

We also conduct human evaluations with Amazon Mechanical Turk.
For 300 generated sequences from each dataset, we show one video from our model and one video from a baseline model for the same test data.
The workers are asked to mark their preferences on ours versus the baseline model on visual qulity and action consistency (Fig~\ref{fig:human_evaluation_all}).

\textbf{Video Quality:}
Tab~\ref{table:fvd} shows the result on Fr\'{e}chet Video Distance (FVD)~\cite{unterthiner2018towards}.
FVD measures the distance between the distributions of the ground truth and generated video sequences.
FVD is an extension of FID~\cite{heusel2017gans} for videos and is suitable for measuring the quality of generated videos.
{Our model} achieves lower FVD than all baseline models except for GameGAN on Gibson.
The primary reason we suspect is that our model on Gibson sometimes slightly changes the brightness. In contrast, GameGAN, being a model directly learned on pixel space, produced more consistent brightness.
Human evaluation of visual quality (Fig~\ref{fig:human_evaluation_all})  shows that subjects strongly prefer our model, even for Gibson.

%
%

\begin{table}[h!]
\vspace{-3mm}
\centering
\begin{small}
\begin{tabular}{c | c c c}
\hline\hline
 & \multicolumn{3}{c}{Frechet Video Distance $\downarrow$}  \\
Model & Carla & Gibson & RWD \\ [0.5ex]

\hline

Action-RNN & 1523.3 & 1109.2 & 2560.7 \\
World Model & 1663.0 & 1212.0 & 2795.6 \\
World Model* & 1138.6 & 561.1 & 591.7 \\
SAVP & 1018.2 & 470.7 & 977.9 \\
GameGAN & 739.5 & \textbf{311.4} & 801.0 \\
Ours & \textbf{281.9} & 360.0 & \textbf{518.0} \\ [1ex]
\hline
\end{tabular}
\end{small}
\vspace{-2mm}
\caption{
  \small   Results on FVD~\cite{unterthiner2018towards}. Lower is better.
}
\label{table:fvd}
\vspace{-3mm}
\end{table}

\textbf{Action Consistency:}
We measure if generated sequences conform to the input action sequences.
We train a CNN model that takes two images from real videos as input and predicts the action that caused the transition between them.
The model is trained by reducing the mean-squared error loss between the predicted and input actions.
The trained model can be applied to the generated sequences from simulator models to evaluate action consistency.
Table~\ref{table:action_prediction} and human evaluation (Fig~\ref{fig:human_evaluation_all}) show that our model achieves the best performance on all datasets.

\begin{table}[h!]
\vspace{-2mm}
\centering
\begin{small}
\begin{tabular}{c | c c c}
\hline\hline
 & \multicolumn{3}{c}{Action Prediction Loss $\downarrow$}  \\
Model & Carla & Gibson & RWD \\ [0.5ex]

\hline

Action-RNN & 4.850 & 0.062 & 0.586 \\
World Model & 5.310 &  0.167 & 0.721 \\
World Model* & 17.384 & 0.082 & 0.885 \\
SAVP & 3.178 & 0.070 & 0.645 \\
GameGAN & 2.341 & 0.065 & 0.638 \\
Ours & \textbf{1.686} & \textbf{0.045} & \textbf{0.412} \\ [1ex]
\hline \hline
Real Data & 0.370 & 0.005 & 0.159 \\ [1ex]
\hline
\end{tabular}
\end{small}
\vspace{-2mm}
\caption{
    \small Results on Action Prediction. Lower is better.
}
\label{table:action_prediction}
\vspace{-2mm}
\end{table}

\begin{figure}[!h]
    \vspace{-2mm}
    \begin{center}
        \includegraphics[width=0.95\linewidth]{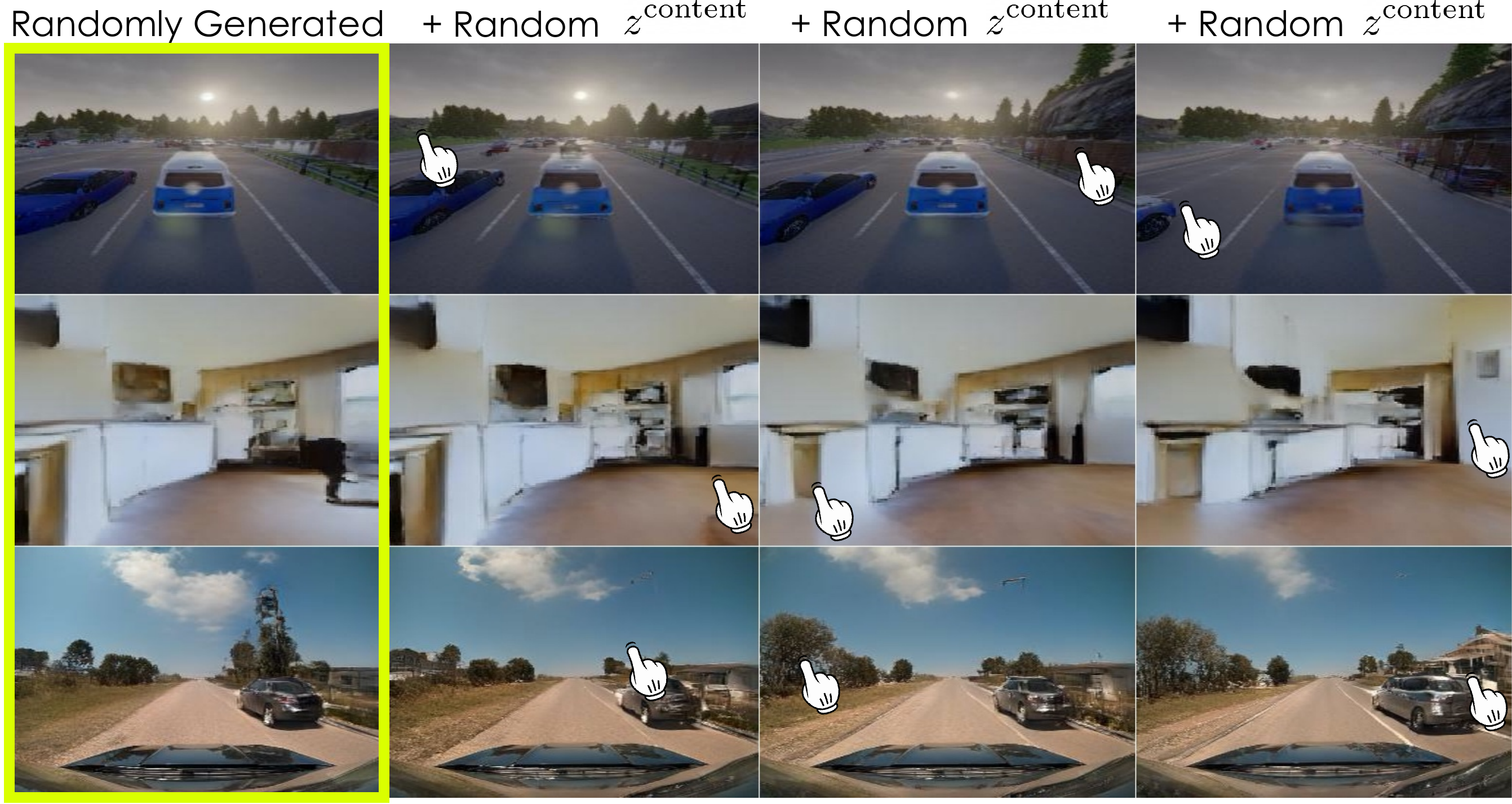}
    \end{center}
    \vspace{-5.5mm}
    \caption[]{
   \small Users can randomly sample a vector for a grid cell in $z^{theme}$ to change the cell's content.
    The white figner corresponds to the locations a user clicked to modify.
    }
    \label{fig:change_part}
    \vspace{-1.5mm}
\end{figure}

\begin{figure}[!h]
    \vspace{-2mm}
    \begin{center}
        \includegraphics[width=0.95\linewidth]{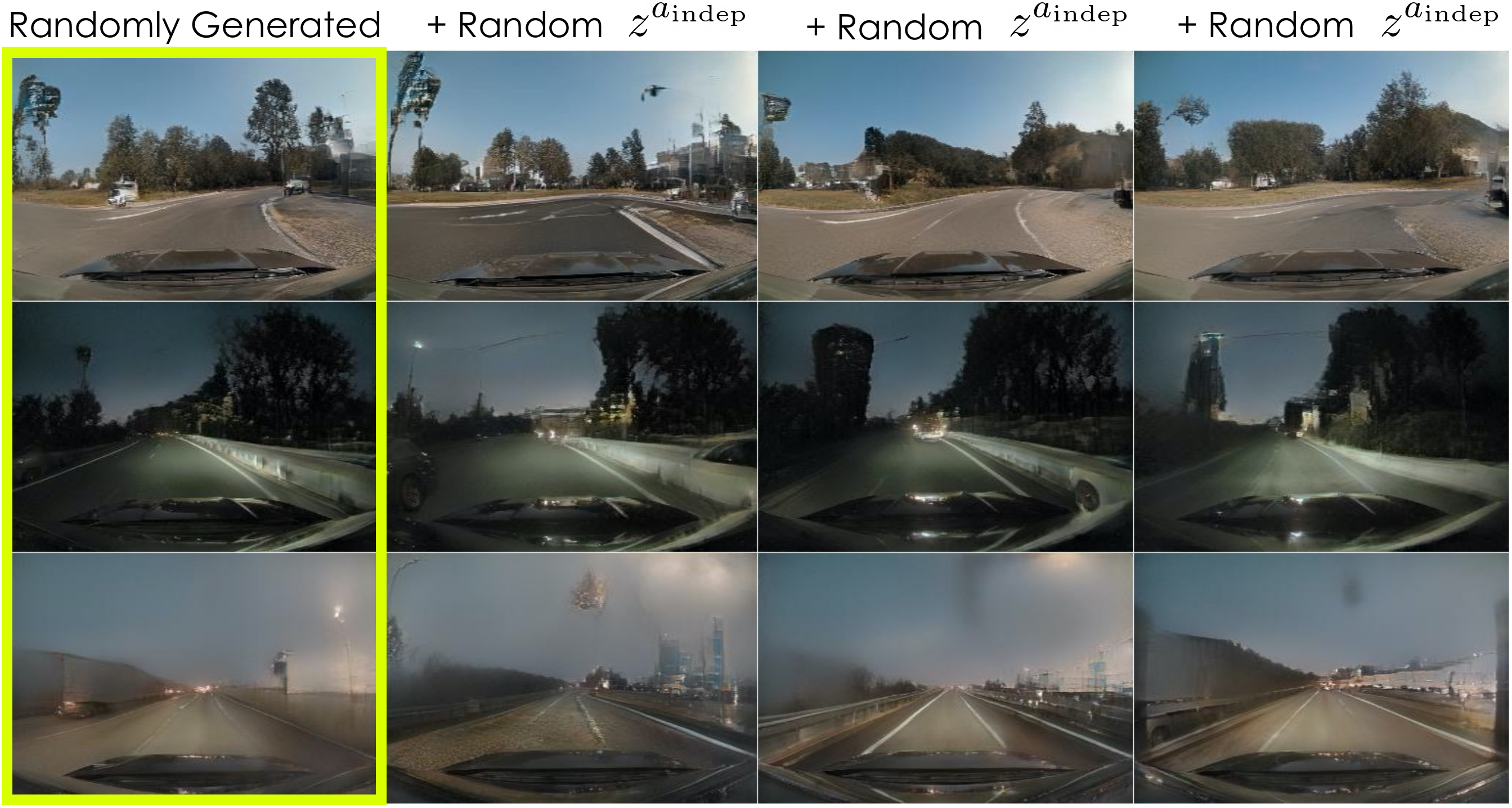}
    \end{center}
    \vspace{-5.5mm}
    \caption[]{
    \small Swapping $z^{a_{\text{indep}}}$ modifies objects in a scene while keeping layout, such as the shape of the road, consistent.
    \textbf{Top:} right turn,
    \textbf{Middle:} road for slight left,
    \textbf{Bottom:} straight road.
    }
    \label{fig:random_aindep}
    \vspace{-4mm}
\end{figure}

\begin{figure}[!h]
    \vspace{-2mm}
    \begin{center}
        \includegraphics[width=\linewidth]{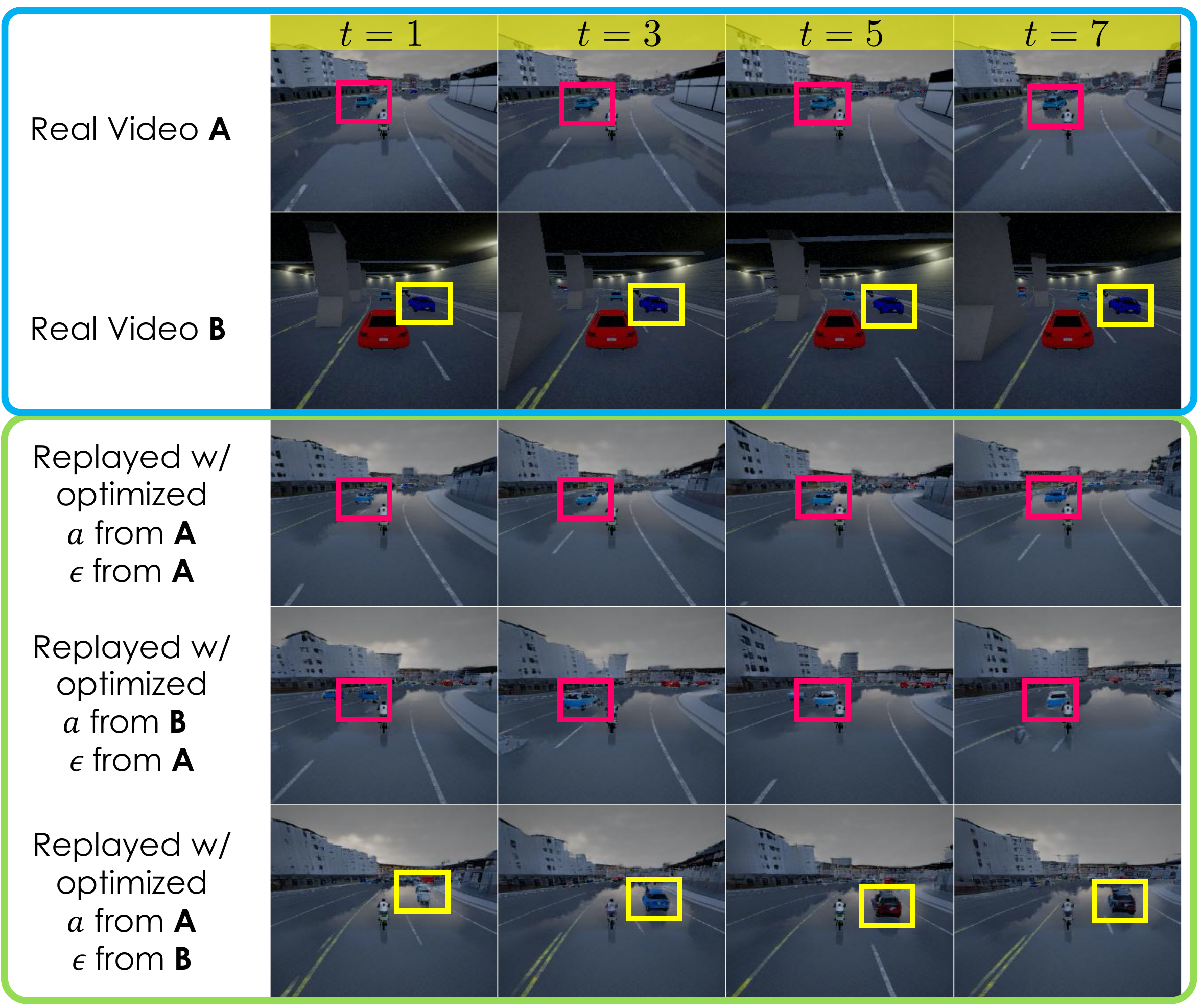}
    \end{center}
    \vspace{-6mm}
    \caption[]{
    \small We optimize action ($a_{0..T-1}^A, a_{0..T-1}^B$) and stochastic variable sequences ($\epsilon_{0..T-1}^A, \epsilon_{0..T-1}^B$) for real videos A and B.
    Let $z_0^A$ be the latent code of A's initial frame.
    We show re-played sequences using
    ($z_0^A, a^A, \epsilon^A$),
    ($z_0^A, a^B, \epsilon^A$) and
    ($z_0^A, a^A, \epsilon^B$).
    }
    \label{fig:diffSim_epsilon_swap}
    \vspace{-5mm}
\end{figure}

\begin{figure*}[!h]
\vspace{-1mm}
    \begin{center}
        \includegraphics[width=0.82\textwidth]{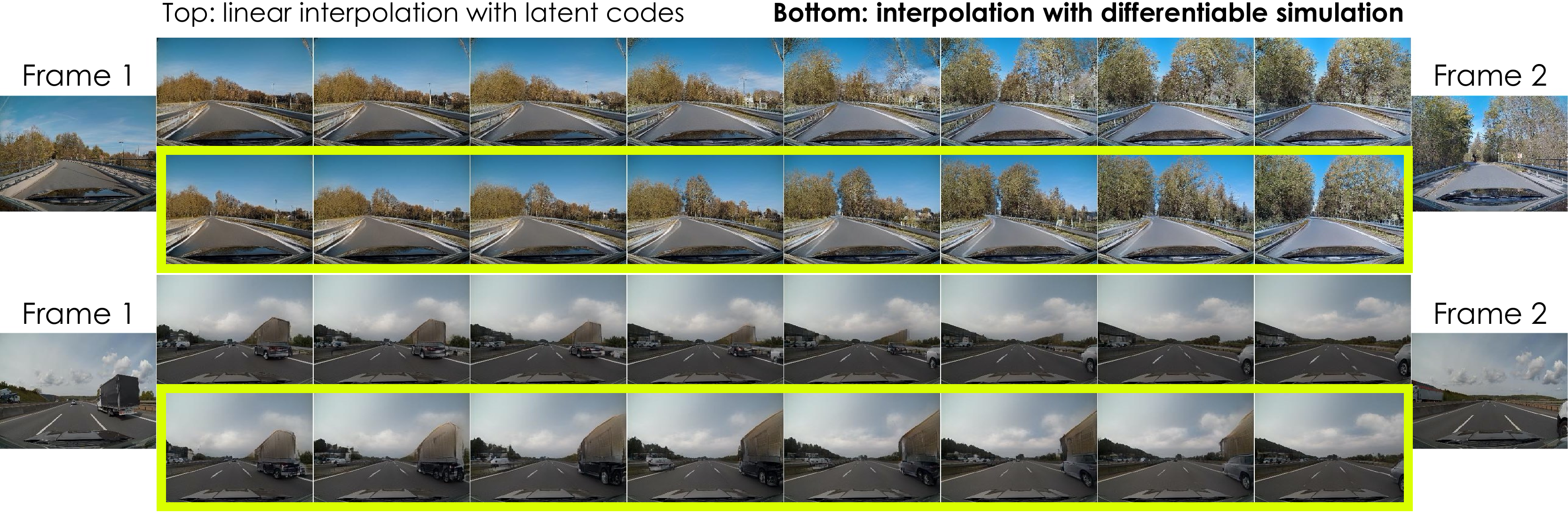}
    \end{center}
    \vspace{-6.5mm}
    \caption[]{
       \small  \textbf{Frame Interpolation}
        We run differentiable simulation to produce Frame 2 given Frame 1.
        \textbf{Top:} Linear interpolation in latent space does not account for transition dynamics correctly.
        \textbf{Bottom:} {\Name} keeps dynamics consistent with respect to the environment.
    }
    \label{fig:interpolation}
     \vspace{-2.5mm}
\end{figure*}

\subsection{Controllability and Differentiable Simulation}
\label{sec:qual_results}
{\Name} learns to disentangle factors comprising a scene without supervision, and it naturally allows controllability on all $z$s as $z^{a_{\text{dep}}}$, $z^{a_{\text{indep}}}$, $z^{\text{content}}$ and $z^{\text{theme}}$ can be sampled from the prior distribution.
Fig~\ref{fig:change_theme} demonstrates how we can change the background color or weather condition by sampling and swapping $z^{\text{theme}}$.
Fig~\ref{fig:random_aindep} shows how sampling different $z^{a_{\text{indep}}}$ modifies the interior parts, such as object shapes, while keeping the layout and theme consistent.
This allows users to sample various scenarios for specific layout shapes.
As $z^{\text{content}}$ is a spatial tensor, we can sample each grid cell to change the content of the cell.
In the bottom row of Fig~\ref{fig:change_part}, a user clicks specific locations to erase a tree, add a tree, and add a building.

We also record the sampled $z$s corresponding to specific content and build an editable neural simulator, as in Fig~\ref{fig:overview}.
This editing procedure lets users create unique simulation scenarios and selectively focus on the ones they want.
Note that we can even sample the first screen, unlike some previous works such as GameGAN~\cite{kim2020learning}.

\begin{figure}[!h]
    \vspace{-2mm}
    \begin{center}
        \includegraphics[width=0.95\linewidth]{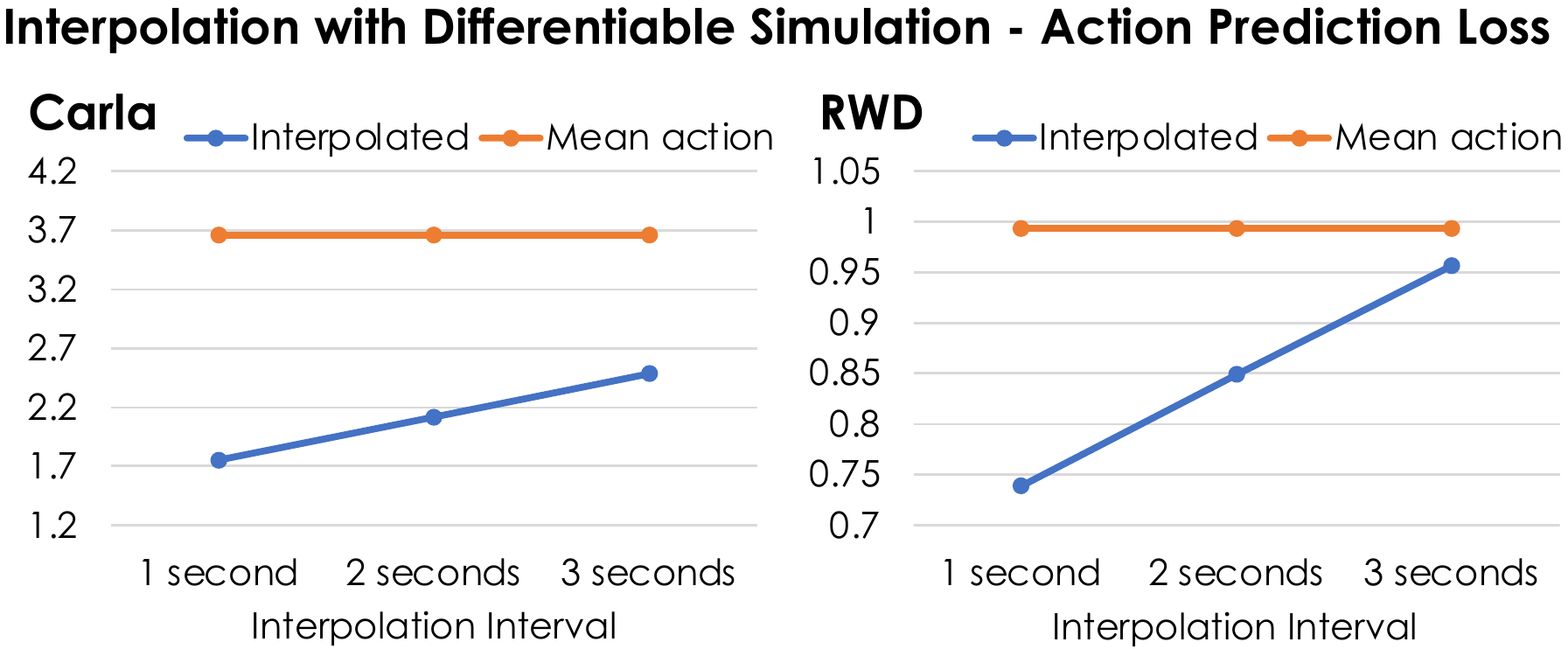}
    \end{center}
    \vspace{-5.5mm}
    \caption[]{
       \small Optimized actions from frame interpolation discovers in-between actions.
        Mean action measures action prediction loss when the mean of actions from the training dataset is used as  input.
    }
    \label{fig:interpolation_action_prediction}
    \vspace{-4.5mm}
\end{figure}

\textbf{Differentiable Simulation:}
Sec~\ref{sec:diff_sim} introduces how we can create an editable simulation environment from a real video by recovering the underlying actions $a$ and stochastic variables $\epsilon$ with Eq.(\ref{eq:diffSim}).
Fig~\ref{fig:diffSim_epsilon_swap} illustrates the result of differentiable simulation.
The third row exhibits how we can recover the original video A by running {\Name} with optimized $a$ and $\epsilon$.
To verify we have recovered $a$ successfully and not just overfitted using $\epsilon$, we evaluate the quality of optimized $a$ from test data using the Action Prediction loss from Tab~\ref{table:action_prediction}.
Optimized $a$ results in a loss of 1.91 and 0.57 for Carla and RWD, respectively.
These numbers are comparable to Tab~\ref{table:action_prediction} and much lower than the baseline performances of 3.64 and 1.01, calculated with the mean of actions from the training data, demonstrating that {\Name} can recover unobserved actions successfully.
We can even recover $a$ and $\epsilon$ for non-existing intermediate frames.
That is, we can do \textit{frame interpolation} to discover in-between frames given a reference and a future frame.
If the time between the two frames is small, even a naive linear interpolation could work.
However, for a large gap ($\geq$ 1 second), it is necessary to reason about the environment's dynamics to properly interpolate objects in a scene.
We modify Eq.(\ref{eq:diffSim}) to minimize the reconstruction term for the last frame $z_T$ only, and add a regularization $||z_t-z_{t-1}||$ on the intermediate $z$s.
Fig~\ref{fig:interpolation} shows the result.
Top row, which shows interpolation in the latent space, produces reasonable in-between frames, but if inspected closely, we can see the transition is unnatural (\eg a tree appears out of nowhere).
On the contrary, with differentiable simulation, we can see how it learns to utilize the dynamics of {\Name} to produce plausible transitions between frames.
In Fig~\ref{fig:interpolation_action_prediction}, we calculate the action prediction loss with optimized actions from frame interpolation.
We discover optimized actions that follow the ground-truth actions closely when we interpolate frames one second apart.
As the interpolation interval becomes larger, the loss increases since many possible action sequences lead to the same resulting frame.
This shows the possibility of using differentiable simulation for video compression as it can decode missing intermediate frames.

Differentiable simulation also allows replaying the same scenario with different inputs.
In Fig~\ref{fig:diffSim_epsilon_swap}, we get optimized $a^A, \epsilon^A$ and $a^B, \epsilon^B$ for two driving videos, A and B.
We replay starting with the encoded first frame $z_0^A$ of A.
On the fourth row, ran with ($a^B$,$\epsilon^A$), we see that a vehicle is placed at the same location as A, but since we use the slightly-left action sequence $a^B$, the ego agent changes the lane and slides toward the vehicle.
The fifth row, replayed with ($a^A$,$\epsilon^B$), shows the same ego-agent's trajectory as A, but it puts a vehicle at the same location as B due to $\epsilon^B$. This effectively shows that we can \textit{blend-in} two different scenarios together.
Furthermore, we can modify the content and run a simulation with the environment inferred from a video.
In Fig~\ref{fig:diffSim_pilotnet}, we create a simulation environment from a RWD test data, and replay with modified objects and weather.

\subsection{Additional Experiments}

\begin{figure}[!h]
    \vspace{-2mm}
    \begin{center}
        \includegraphics[width=0.85\linewidth]{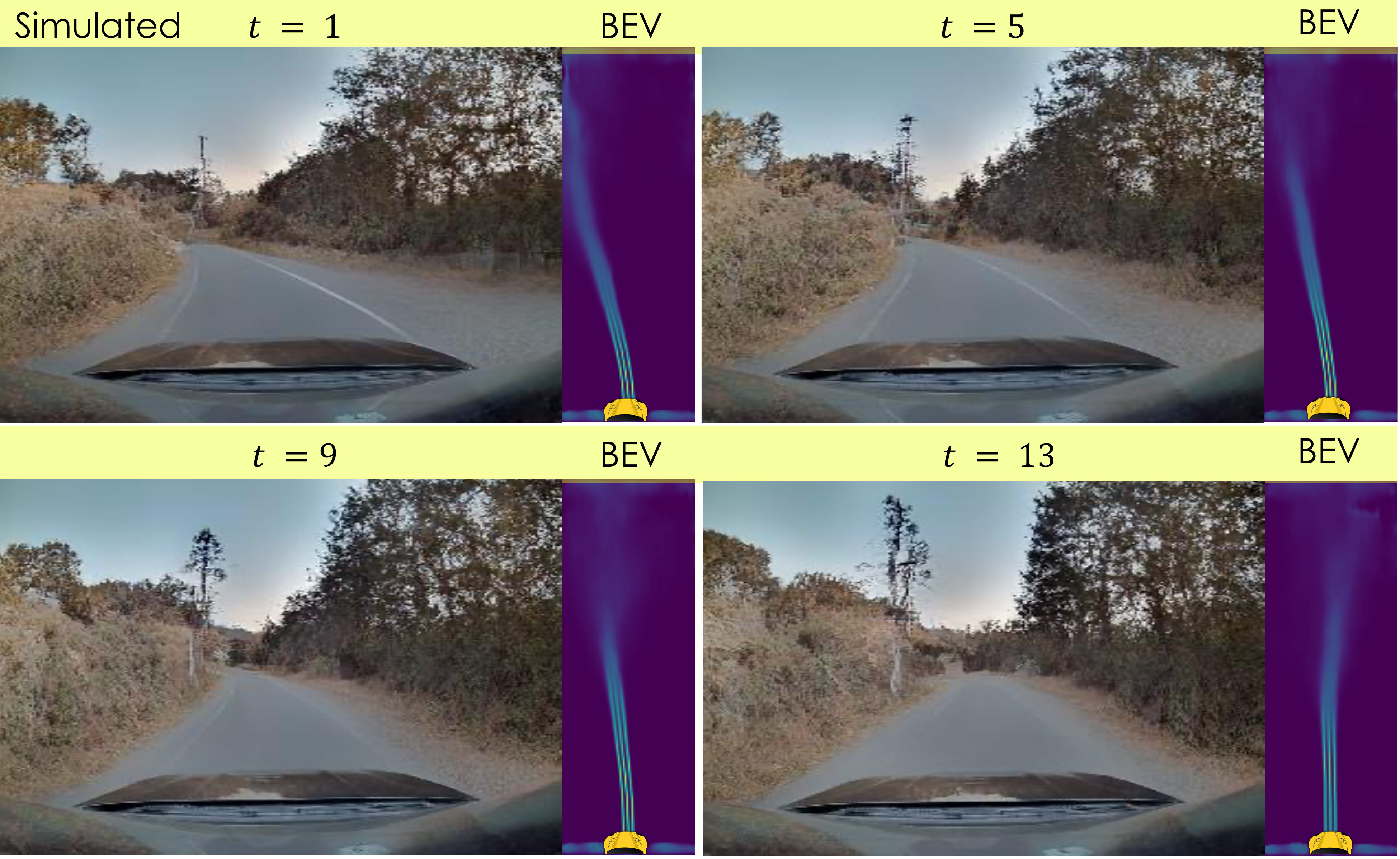}
    \end{center}
    \vspace{-5.5mm}
    \caption[]{
       \small Bird's-Eve-View (BEV) lane prediction with LiftSplat~\cite{philion2020lift} model on generated scenes.
    }
    \label{fig:liftsplat_bev}
    \vspace{-3mm}
\end{figure}

\textbf{LiftSplat}
\cite{philion2020lift} proposed a model for producing the Bird's-Eye-View (BEV) representation of a scene from camera images.
We use LiftSplat to get BEV lane predictions from a simulated sequence from {\Name} (Fig~\ref{fig:liftsplat_bev}).
Simulated scenes are realistic enough for LiftSplat to produce accurate predictions.
This shows the potential of {\Name} being used with other perception models to be useful for downstream tasks such as training an autonomous driving agent.
Furthermore, in real-time driving, LiftSplat can potentially employ {\Name}'s simulated frames as a safety measure to be robust to sudden camera drop-outs.

\textbf{Plain StyleGAN latent space:}
StyleGAN~\cite{karras2020analyzing} proposes an optimization scheme to project images into their latent codes without an encoder.
The projection process optimizes each image and requires significant time ($\sim$19200 GPU hours for Gibson).
Therefore, we use 25\% of Gibson data to compare with the projection approach.
We train the same dynamics model on top of the projected and proposed latent spaces.
The projection approach resulted in FVD of \textbf{636.8} with the action prediction loss of \textbf{0.225}, whereas
ours achieved \textbf{411.9} (FVD) and \textbf{0.050} (action prediction loss).


\section{Conclusion}
\label{sec:conclusion}
We proposed {\Name} for a controllable high-quality simulation.
{\Name} leverages a novel encoder and an image GAN to produce a latent space on which the proposed dynamics engine learns the transitions between frames.
{\Name} allows sampling and disentangling of different components of a scene without supervision.
This lets users interactively edit scenes during a simulation and produce unique scenarios.
We showcased \textit{differentiable simulation} which opens up promising ways for utilizing real-world videos to discover the underlying factors of variations and train robots in the re-created environments.

% ...

\clearpage

{\small
\bibliographystyle{ieee_fullname}
\bibliography{egbib}

\begin{thebibliography}{10}\itemsep=-1pt

\bibitem{DBLP:journals/corr/abs-1710-11252}
Mohammad Babaeizadeh, Chelsea Finn, Dumitru Erhan, Roy~H. Campbell, and Sergey
  Levine.
\newblock Stochastic variational video prediction.
\newblock {\em CoRR}, abs/1710.11252, 2017.

\bibitem{bellemare2013arcade}
Marc~G Bellemare, Yavar Naddaf, Joel Veness, and Michael Bowling.
\newblock The arcade learning environment: An evaluation platform for general
  agents.
\newblock {\em Journal of Artificial Intelligence Research}, 47:253--279, 2013.

\bibitem{chiappa2017recurrent}
Silvia Chiappa, S{\'e}bastien Racaniere, Daan Wierstra, and Shakir Mohamed.
\newblock Recurrent environment simulators.
\newblock {\em arXiv preprint arXiv:1704.02254}, 2017.

\bibitem{chung2014empirical}
Junyoung Chung, Caglar Gulcehre, KyungHyun Cho, and Yoshua Bengio.
\newblock Empirical evaluation of gated recurrent neural networks on sequence
  modeling.
\newblock {\em arXiv preprint arXiv:1412.3555}, 2014.

\bibitem{dvdgan}
Aidan Clark, Jeff Donahue, and Karen Simonyan.
\newblock Efficient video generation on complex datasets.
\newblock {\em CoRR}, abs/1907.06571, 2019.

\bibitem{deisenroth2011pilco}
Marc Deisenroth and Carl~E Rasmussen.
\newblock Pilco: A model-based and data-efficient approach to policy search.
\newblock In {\em Proceedings of the 28th International Conference on machine
  learning (ICML-11)}, pages 465--472, 2011.

\bibitem{DBLP:journals/corr/abs-1802-07687}
Emily Denton and Rob Fergus.
\newblock Stochastic video generation with a learned prior.
\newblock {\em CoRR}, abs/1802.07687, 2018.

\bibitem{denton2017unsupervised}
Emily~L Denton et~al.
\newblock Unsupervised learning of disentangled representations from video.
\newblock In {\em Advances in neural information processing systems}, pages
  4414--4423, 2017.

\bibitem{devaranjan2020meta}
Jeevan Devaranjan, Amlan Kar, and Sanja Fidler.
\newblock Meta-sim2: Unsupervised learning of scene structure for synthetic
  data generation.
\newblock {\em arXiv preprint arXiv:2008.09092}, 2020.

\bibitem{Dosovitskiy17}
Alexey Dosovitskiy, German Ros, Felipe Codevilla, Antonio Lopez, and Vladlen
  Koltun.
\newblock {CARLA}: {An} open urban driving simulator.
\newblock In {\em Proceedings of the 1st Annual Conference on Robot Learning},
  pages 1--16, 2017.

\bibitem{dumoulin2016learned}
Vincent Dumoulin, Jonathon Shlens, and Manjunath Kudlur.
\newblock A learned representation for artistic style.
\newblock {\em arXiv preprint arXiv:1610.07629}, 2016.

\bibitem{finn2016unsupervised}
Chelsea Finn, Ian Goodfellow, and Sergey Levine.
\newblock Unsupervised learning for physical interaction through video
  prediction.
\newblock In {\em Advances in neural information processing systems}, pages
  64--72, 2016.

\bibitem{ghiasi2017exploring}
Golnaz Ghiasi, Honglak Lee, Manjunath Kudlur, Vincent Dumoulin, and Jonathon
  Shlens.
\newblock Exploring the structure of a real-time, arbitrary neural artistic
  stylization network.
\newblock {\em arXiv preprint arXiv:1705.06830}, 2017.

\bibitem{gan}
Ian~J. Goodfellow, Jean Pouget-Abadie, Mehdi Mirza, Bing Xu, David
  Warde-Farley, Sherjil Ozair, Aaron Courville, and Yoshua Bengio.
\newblock Generative adversarial networks, 2014.

\bibitem{ha2018recurrent}
David Ha and J{\"u}rgen Schmidhuber.
\newblock Recurrent world models facilitate policy evolution.
\newblock In {\em Advances in Neural Information Processing Systems}, pages
  2450--2462, 2018.

\bibitem{hafner2019learning}
Danijar Hafner, Timothy Lillicrap, Ian Fischer, Ruben Villegas, David Ha,
  Honglak Lee, and James Davidson.
\newblock Learning latent dynamics for planning from pixels.
\newblock In {\em International Conference on Machine Learning}, pages
  2555--2565. PMLR, 2019.

\bibitem{he2016deep}
Kaiming He, Xiangyu Zhang, Shaoqing Ren, and Jian Sun.
\newblock Deep residual learning for image recognition.
\newblock In {\em Proceedings of the IEEE conference on computer vision and
  pattern recognition}, pages 770--778, 2016.

\bibitem{heusel2017gans}
Martin Heusel, Hubert Ramsauer, Thomas Unterthiner, Bernhard Nessler, and Sepp
  Hochreiter.
\newblock Gans trained by a two time-scale update rule converge to a local nash
  equilibrium.
\newblock In {\em Advances in neural information processing systems}, pages
  6626--6637, 2017.

\bibitem{higgins2016beta}
Irina Higgins, Loic Matthey, Arka Pal, Christopher Burgess, Xavier Glorot,
  Matthew Botvinick, Shakir Mohamed, and Alexander Lerchner.
\newblock beta-vae: Learning basic visual concepts with a constrained
  variational framework.
\newblock 2016.

\bibitem{hochreiter1997long}
Sepp Hochreiter and J{\"u}rgen Schmidhuber.
\newblock Long short-term memory.
\newblock {\em Neural computation}, 9(8):1735--1780, 1997.

\bibitem{DBLP:journals/corr/abs-1806-04166}
Jun{-}Ting Hsieh, Bingbin Liu, De{-}An Huang, Li Fei{-}Fei, and Juan~Carlos
  Niebles.
\newblock Learning to decompose and disentangle representations for video
  prediction.
\newblock {\em CoRR}, abs/1806.04166, 2018.

\bibitem{huang2017arbitrary}
Xun Huang and Serge Belongie.
\newblock Arbitrary style transfer in real-time with adaptive instance
  normalization.
\newblock In {\em Proceedings of the IEEE International Conference on Computer
  Vision}, pages 1501--1510, 2017.

\bibitem{ioffe2015batch}
Sergey Ioffe and Christian Szegedy.
\newblock Batch normalization: Accelerating deep network training by reducing
  internal covariate shift.
\newblock {\em arXiv preprint arXiv:1502.03167}, 2015.

\bibitem{isola2017image}
Phillip Isola, Jun-Yan Zhu, Tinghui Zhou, and Alexei~A Efros.
\newblock Image-to-image translation with conditional adversarial networks.
\newblock In {\em Proceedings of the IEEE conference on computer vision and
  pattern recognition}, pages 1125--1134, 2017.

\bibitem{kaiser2019model}
Lukasz Kaiser, Mohammad Babaeizadeh, Piotr Milos, Blazej Osinski, Roy~H
  Campbell, Konrad Czechowski, Dumitru Erhan, Chelsea Finn, Piotr Kozakowski,
  Sergey Levine, et~al.
\newblock Model-based reinforcement learning for atari.
\newblock {\em arXiv preprint arXiv:1903.00374}, 2019.

\bibitem{DBLP:journals/corr/KalchbrennerOSD16}
Nal Kalchbrenner, A{\"{a}}ron van~den Oord, Karen Simonyan, Ivo Danihelka,
  Oriol Vinyals, Alex Graves, and Koray Kavukcuoglu.
\newblock Video pixel networks.
\newblock {\em CoRR}, abs/1610.00527, 2016.

\bibitem{kar2019meta}
Amlan Kar, Aayush Prakash, Ming-Yu Liu, Eric Cameracci, Justin Yuan, Matt
  Rusiniak, David Acuna, Antonio Torralba, and Sanja Fidler.
\newblock Meta-sim: Learning to generate synthetic datasets.
\newblock In {\em Proceedings of the IEEE International Conference on Computer
  Vision}, pages 4551--4560, 2019.

\bibitem{karras2019style}
Tero Karras, Samuli Laine, and Timo Aila.
\newblock A style-based generator architecture for generative adversarial
  networks.
\newblock In {\em Proceedings of the IEEE conference on computer vision and
  pattern recognition}, pages 4401--4410, 2019.

\bibitem{karras2020analyzing}
Tero Karras, Samuli Laine, Miika Aittala, Janne Hellsten, Jaakko Lehtinen, and
  Timo Aila.
\newblock Analyzing and improving the image quality of stylegan.
\newblock In {\em Proceedings of the IEEE/CVF Conference on Computer Vision and
  Pattern Recognition}, pages 8110--8119, 2020.

\bibitem{kim2020learning}
Seung~Wook Kim, Yuhao Zhou, Jonah Philion, Antonio Torralba, and Sanja Fidler.
\newblock Learning to simulate dynamic environments with gamegan.
\newblock In {\em Proceedings of the IEEE/CVF Conference on Computer Vision and
  Pattern Recognition}, pages 1231--1240, 2020.

\bibitem{kim2014convolutional}
Yoon Kim.
\newblock Convolutional neural networks for sentence classification.
\newblock {\em arXiv preprint arXiv:1408.5882}, 2014.

\bibitem{kingma2014adam}
Diederik~P Kingma and Jimmy Ba.
\newblock Adam: A method for stochastic optimization.
\newblock {\em arXiv preprint arXiv:1412.6980}, 2014.

\bibitem{kingma2014autoencoding}
Diederik~P Kingma and Max Welling.
\newblock Auto-encoding variational bayes, 2014.

\bibitem{THOR}
Eric Kolve, Roozbeh Mottaghi, Daniel Gordon, Yuke Zhu, Abhinav Gupta, and Ali
  Farhadi.
\newblock Ai2-thor: An interactive 3d environment for visual ai.
\newblock In {\em arXiv:1712.05474}, 2017.

\bibitem{DBLP:journals/corr/abs-1903-01434}
Manoj Kumar, Mohammad Babaeizadeh, Dumitru Erhan, Chelsea Finn, Sergey Levine,
  Laurent Dinh, and Durk Kingma.
\newblock Videoflow: {A} flow-based generative model for video.
\newblock {\em CoRR}, abs/1903.01434, 2019.

\bibitem{DBLP:journals/corr/LarsenSW15}
Anders Boesen~Lindbo Larsen, S{\o}ren~Kaae S{\o}nderby, and Ole Winther.
\newblock Autoencoding beyond pixels using a learned similarity metric.
\newblock {\em CoRR}, abs/1512.09300, 2015.

\bibitem{lee2018stochastic}
Alex~X Lee, Richard Zhang, Frederik Ebert, Pieter Abbeel, Chelsea Finn, and
  Sergey Levine.
\newblock Stochastic adversarial video prediction.
\newblock {\em arXiv preprint arXiv:1804.01523}, 2018.

\bibitem{lim2017geometric}
Jae~Hyun Lim and Jong~Chul Ye.
\newblock Geometric gan.
\newblock {\em arXiv preprint arXiv:1705.02894}, 2017.

\bibitem{lotter2016deep}
William Lotter, Gabriel Kreiman, and David Cox.
\newblock Deep predictive coding networks for video prediction and unsupervised
  learning.
\newblock {\em arXiv preprint arXiv:1605.08104}, 2016.

\bibitem{maas2013rectifier}
Andrew~L Maas, Awni~Y Hannun, and Andrew~Y Ng.
\newblock Rectifier nonlinearities improve neural network acoustic models.
\newblock In {\em Proc. icml}, volume~30, page~3, 2013.

\bibitem{mallya2020worldconsistent}
Arun Mallya, Ting-Chun Wang, Karan Sapra, and Ming-Yu Liu.
\newblock World-consistent video-to-video synthesis, 2020.

\bibitem{manivasagam2020lidarsim}
Sivabalan Manivasagam, Shenlong Wang, Kelvin Wong, Wenyuan Zeng, Mikita
  Sazanovich, Shuhan Tan, Bin Yang, Wei-Chiu Ma, and Raquel Urtasun.
\newblock Lidarsim: Realistic lidar simulation by leveraging the real world.
\newblock In {\em Proceedings of the IEEE/CVF Conference on Computer Vision and
  Pattern Recognition}, pages 11167--11176, 2020.

\bibitem{mathieu2016deep}
Michael Mathieu, Camille Couprie, and Yann LeCun.
\newblock Deep multi-scale video prediction beyond mean square error, 2016.

\bibitem{mescheder2018training}
Lars Mescheder, Andreas Geiger, and Sebastian Nowozin.
\newblock Which training methods for gans do actually converge?
\newblock {\em arXiv preprint arXiv:1801.04406}, 2018.

\bibitem{minderer2019unsupervised}
Matthias Minderer, Chen Sun, Ruben Villegas, Forrester Cole, Kevin~P Murphy,
  and Honglak Lee.
\newblock Unsupervised learning of object structure and dynamics from videos.
\newblock In {\em Advances in Neural Information Processing Systems}, pages
  92--102, 2019.

\bibitem{miyato2018spectral}
Takeru Miyato, Toshiki Kataoka, Masanori Koyama, and Yuichi Yoshida.
\newblock Spectral normalization for generative adversarial networks.
\newblock {\em arXiv preprint arXiv:1802.05957}, 2018.

\bibitem{oh2015action}
Junhyuk Oh, Xiaoxiao Guo, Honglak Lee, Richard~L Lewis, and Satinder Singh.
\newblock Action-conditional video prediction using deep networks in atari
  games.
\newblock In {\em Advances in neural information processing systems}, pages
  2863--2871, 2015.

\bibitem{philion2020lift}
Jonah Philion and Sanja Fidler.
\newblock Lift, splat, shoot: Encoding images from arbitrary camera rigs by
  implicitly unprojecting to 3d.
\newblock {\em arXiv preprint arXiv:2008.05711}, 2020.

\bibitem{VirtualHome2018}
Xavier Puig, Kevin Ra, Marko Boben, Jiaman Li, Tingwu Wang, Sanja Fidler, and
  Antonio Torralba.
\newblock Virtualhome: Simulating household activities via programs.
\newblock In {\em CVPR}, 2018.

\bibitem{DBLP:journals/corr/RanzatoSBMCC14}
Marc'Aurelio Ranzato, Arthur Szlam, Joan Bruna, Micha{\"{e}}l Mathieu, Ronan
  Collobert, and Sumit Chopra.
\newblock Video (language) modeling: a baseline for generative models of
  natural videos.
\newblock {\em CoRR}, abs/1412.6604, 2014.

\bibitem{ruiz2018learning}
Nataniel Ruiz, Samuel Schulter, and Manmohan Chandraker.
\newblock Learning to simulate.
\newblock {\em arXiv preprint arXiv:1810.02513}, 2018.

\bibitem{saito2017temporal}
Masaki Saito, Eiichi Matsumoto, and Shunta Saito.
\newblock Temporal generative adversarial nets with singular value clipping,
  2017.

\bibitem{Saito2018TGANv2ET}
M. Saito and Shunta Saito.
\newblock Tganv2: Efficient training of large models for video generation with
  multiple subsampling layers.
\newblock {\em ArXiv}, abs/1811.09245, 2018.

\bibitem{AirSim}
Shital Shah, Debadeepta Dey, Chris Lovett, and Ashish Kapoor.
\newblock {A}erial {I}nformatics and {R}obotics platform.
\newblock Technical Report {M}{S}{R}-{T}{R}-2017-9, Microsoft Research, 2017.

\bibitem{shaham2019singan}
Tamar~Rott Shaham, Tali Dekel, and Tomer Michaeli.
\newblock Singan: Learning a generative model from a single natural image.
\newblock In {\em Proceedings of the IEEE International Conference on Computer
  Vision}, pages 4570--4580, 2019.

\bibitem{DBLP:journals/corr/SrivastavaMS15}
Nitish Srivastava, Elman Mansimov, and Ruslan Salakhutdinov.
\newblock Unsupervised learning of video representations using lstms.
\newblock {\em CoRR}, abs/1502.04681, 2015.

\bibitem{sutton1990integrated}
Richard~S Sutton.
\newblock Integrated architectures for learning, planning, and reacting based
  on approximating dynamic programming.
\newblock In {\em Machine learning proceedings 1990}, pages 216--224. Elsevier,
  1990.

\bibitem{todorov2012mujoco}
Emanuel Todorov, Tom Erez, and Yuval Tassa.
\newblock Mujoco: A physics engine for model-based control.
\newblock In {\em 2012 IEEE/RSJ International Conference on Intelligent Robots
  and Systems}, pages 5026--5033. IEEE, 2012.

\bibitem{tran2017hierarchical}
Dustin Tran, Rajesh Ranganath, and David Blei.
\newblock Hierarchical implicit models and likelihood-free variational
  inference.
\newblock In {\em Advances in Neural Information Processing Systems}, pages
  5523--5533, 2017.

\bibitem{mocogan}
Sergey Tulyakov, Ming{-}Yu Liu, Xiaodong Yang, and Jan Kautz.
\newblock Mocogan: Decomposing motion and content for video generation.
\newblock {\em CoRR}, abs/1707.04993, 2017.

\bibitem{unterthiner2018towards}
Thomas Unterthiner, Sjoerd van Steenkiste, Karol Kurach, Raphael Marinier,
  Marcin Michalski, and Sylvain Gelly.
\newblock Towards accurate generative models of video: A new metric \&
  challenges.
\newblock {\em arXiv preprint arXiv:1812.01717}, 2018.

\bibitem{vondrick2016generating}
Carl Vondrick, Hamed Pirsiavash, and Antonio Torralba.
\newblock Generating videos with scene dynamics, 2016.

\bibitem{vid2vid}
Ting{-}Chun Wang, Ming{-}Yu Liu, Jun{-}Yan Zhu, Guilin Liu, Andrew Tao, Jan
  Kautz, and Bryan Catanzaro.
\newblock Video-to-video synthesis.
\newblock {\em CoRR}, abs/1808.06601, 2018.

\bibitem{DBLP:journals/corr/abs-1906-02634}
Dirk Weissenborn, Oscar T{\"{a}}ckstr{\"{o}}m, and Jakob Uszkoreit.
\newblock Scaling autoregressive video models.
\newblock {\em CoRR}, abs/1906.02634, 2019.

\bibitem{xiazamirhe2018gibsonenv}
Fei Xia, Amir R.~Zamir, Zhi-Yang He, Alexander Sax, Jitendra Malik, and Silvio
  Savarese.
\newblock Gibson env: real-world perception for embodied agents.
\newblock In {\em Computer Vision and Pattern Recognition (CVPR), 2018 IEEE
  Conference on}. IEEE, 2018.

\bibitem{crevnet20}
Wei Yu, Yichao Lu, Steve Easterbrook, and Sanja Fidler.
\newblock Efficient and information-preserving future frame prediction and
  beyond.
\newblock In {\em ICLR}, 2020.

\bibitem{zhang2018unreasonable}
Richard Zhang, Phillip Isola, Alexei~A Efros, Eli Shechtman, and Oliver Wang.
\newblock The unreasonable effectiveness of deep features as a perceptual
  metric.
\newblock In {\em Proceedings of the IEEE conference on computer vision and
  pattern recognition}, pages 586--595, 2018.

\end{thebibliography}


\begin{thebibliography}{10}\itemsep=-1pt

\bibitem{chung2014empirical}
Junyoung Chung, Caglar Gulcehre, KyungHyun Cho, and Yoshua Bengio.
\newblock Empirical evaluation of gated recurrent neural networks on sequence
  modeling.
\newblock {\em arXiv preprint arXiv:1412.3555}, 2014.

\bibitem{gan}
Ian~J. Goodfellow, Jean Pouget-Abadie, Mehdi Mirza, Bing Xu, David
  Warde-Farley, Sherjil Ozair, Aaron Courville, and Yoshua Bengio.
\newblock Generative adversarial networks, 2014.

\bibitem{he2016deep}
Kaiming He, Xiangyu Zhang, Shaoqing Ren, and Jian Sun.
\newblock Deep residual learning for image recognition.
\newblock In {\em Proceedings of the IEEE conference on computer vision and
  pattern recognition}, pages 770--778, 2016.

\bibitem{hochreiter1997long}
Sepp Hochreiter and J{\"u}rgen Schmidhuber.
\newblock Long short-term memory.
\newblock {\em Neural computation}, 9(8):1735--1780, 1997.

\bibitem{ioffe2015batch}
Sergey Ioffe and Christian Szegedy.
\newblock Batch normalization: Accelerating deep network training by reducing
  internal covariate shift.
\newblock {\em arXiv preprint arXiv:1502.03167}, 2015.

\bibitem{isola2017image}
Phillip Isola, Jun-Yan Zhu, Tinghui Zhou, and Alexei~A Efros.
\newblock Image-to-image translation with conditional adversarial networks.
\newblock In {\em Proceedings of the IEEE conference on computer vision and
  pattern recognition}, pages 1125--1134, 2017.

\bibitem{karras2020analyzing}
Tero Karras, Samuli Laine, Miika Aittala, Janne Hellsten, Jaakko Lehtinen, and
  Timo Aila.
\newblock Analyzing and improving the image quality of stylegan.
\newblock In {\em Proceedings of the IEEE/CVF Conference on Computer Vision and
  Pattern Recognition}, pages 8110--8119, 2020.

\bibitem{kingma2014adam}
Diederik~P Kingma and Jimmy Ba.
\newblock Adam: A method for stochastic optimization.
\newblock {\em arXiv preprint arXiv:1412.6980}, 2014.

\bibitem{kingma2014autoencoding}
Diederik~P Kingma and Max Welling.
\newblock Auto-encoding variational bayes, 2014.

\bibitem{lim2017geometric}
Jae~Hyun Lim and Jong~Chul Ye.
\newblock Geometric gan.
\newblock {\em arXiv preprint arXiv:1705.02894}, 2017.

\bibitem{maas2013rectifier}
Andrew~L Maas, Awni~Y Hannun, and Andrew~Y Ng.
\newblock Rectifier nonlinearities improve neural network acoustic models.
\newblock In {\em Proc. icml}, volume~30, page~3, 2013.

\bibitem{mescheder2018training}
Lars Mescheder, Andreas Geiger, and Sebastian Nowozin.
\newblock Which training methods for gans do actually converge?
\newblock {\em arXiv preprint arXiv:1801.04406}, 2018.

\bibitem{miyato2018spectral}
Takeru Miyato, Toshiki Kataoka, Masanori Koyama, and Yuichi Yoshida.
\newblock Spectral normalization for generative adversarial networks.
\newblock {\em arXiv preprint arXiv:1802.05957}, 2018.

\bibitem{philion2020lift}
Jonah Philion and Sanja Fidler.
\newblock Lift, splat, shoot: Encoding images from arbitrary camera rigs by
  implicitly unprojecting to 3d.
\newblock {\em arXiv preprint arXiv:2008.05711}, 2020.

\bibitem{shaham2019singan}
Tamar~Rott Shaham, Tali Dekel, and Tomer Michaeli.
\newblock Singan: Learning a generative model from a single natural image.
\newblock In {\em Proceedings of the IEEE International Conference on Computer
  Vision}, pages 4570--4580, 2019.

\bibitem{tran2017hierarchical}
Dustin Tran, Rajesh Ranganath, and David Blei.
\newblock Hierarchical implicit models and likelihood-free variational
  inference.
\newblock In {\em Advances in Neural Information Processing Systems}, pages
  5523--5533, 2017.

\bibitem{vid2vid}
Ting{-}Chun Wang, Ming{-}Yu Liu, Jun{-}Yan Zhu, Guilin Liu, Andrew Tao, Jan
  Kautz, and Bryan Catanzaro.
\newblock Video-to-video synthesis.
\newblock {\em CoRR}, abs/1808.06601, 2018.

\bibitem{zhang2018unreasonable}
Richard Zhang, Phillip Isola, Alexei~A Efros, Eli Shechtman, and Oliver Wang.
\newblock The unreasonable effectiveness of deep features as a perceptual
  metric.
\newblock In {\em Proceedings of the IEEE conference on computer vision and
  pattern recognition}, pages 586--595, 2018.

\end{thebibliography}
}


\begingroup
\let\clearpage\relax
\onecolumn
\endgroup

\newpage

\begin{large}
\textbf{Supplementary Materials for \\ DriveGAN: Towards a Controllable High-Quality Neural Simulation}
\end{large}
\\
\appendix

\addcontentsline{toc}{section}{Appendices}
\renewcommand{\thesection}{\Alph{section}}

\section{Model Architecture and Training}
We provide detailed descriptions of the model architecture and training process for the pre-trained image encoder-decoder (Sec.~\ref{sec:pretrained}) and dynamics engine (Sec.~\ref{sec:dynamics_engine}).
Unless noted otherwise, we denote tensor dimensions by $H\times W\times D$ where $H$ and $W$ are the spatial height and width of a feature map, and $D$ is the number of channels.

\subsection{Pre-trained Latent Space}
\label{sec:pretrained}
The latent space is pretrained with an encoder, generator and discrminator.
Figure~\ref{fig:model_pretraining} shows the overview of the pretraining model.

\begin{figure}[h!]
    \vspace{-1mm}
    \begin{center}
        \includegraphics[width=0.5\linewidth]{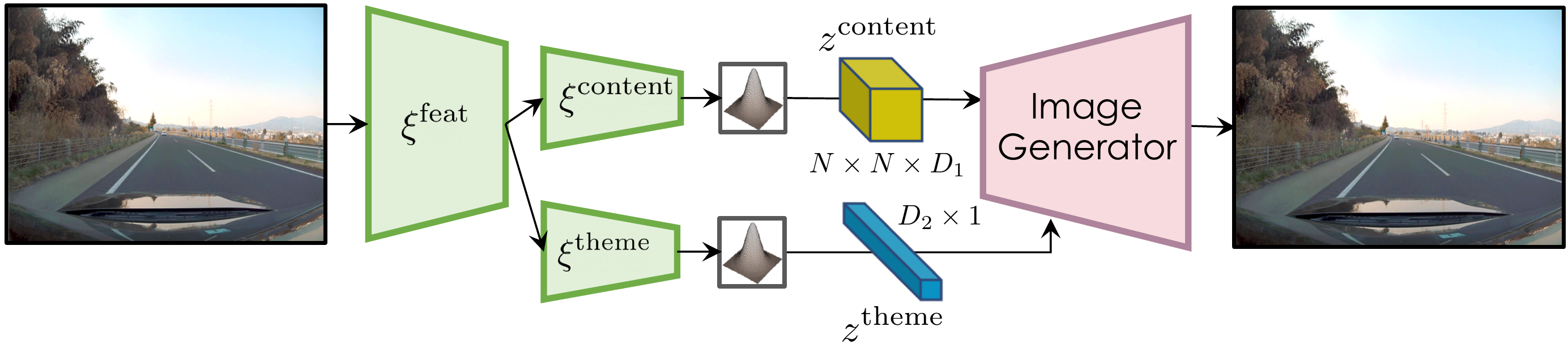}
    \end{center}
    \vspace{-5.5mm}
    \caption[]{
        \small The pretraining stage learns the encoder and decoder for images.
        The encoder $\xi$ produces $z^{\text{content}}$ and $z^{\text{theme}}$ which comprise the disentangled latent space that the dynamics engine trains on.
        The gaussian blocks represent reparameterization steps~\cite{kingma2014autoencoding}.
    }
    \label{fig:model_pretraining}
    \vspace{-2.5mm}
\end{figure}

\subsubsection{Encoder}
Encoder $\xi$ takes an RGB image $x \in \mathbb{R}^{256\times 256\times 3}$ as input and produces disentangled latent codes $z = \{z^{\text{theme}}, z^{\text{content}}\}$ where $z^{\text{theme}} \in \mathbb{R}^{128}$ and $z^{\text{content}} \in \mathbb{R}^{4\times 4\times 64}$.
$\xi$ is composed of a feature extractor $\xi^{\text{feat}}$ and two encoding heads $\xi^{\text{content}}$ and $\xi^{\text{theme}}$.

\begin{table}[h!]
  \begin{center}
    \label{tab:pretrain_encoder}
    \begin{minipage}{.33\linewidth}
        \centering
        \begin{tabular}{c|c}
          \textbf{Layer} & \textbf{Output dimension}\\
          \hline
          Conv2d 3$\times$3 & 256$\times$256$\times$128 \\
          ResBlock   & 128$\times$128$\times$256 \\
          ResBlock   & 64$\times$64$\times$512 \\
          ResBlock   & 32$\times$32$\times$512 \\
          \\

        \end{tabular}
        \caption{$\xi^{\text{feat}}$ architecture}
    \end{minipage}%
    \begin{minipage}{.33\linewidth}
      \centering
      \begin{tabular}{c|c}
        \textbf{Layer} & \textbf{Output dimension}\\
        \hline
        ResBlock   & 16$\times$16$\times$512 \\
        ResBlock   & 8$\times$8$\times$512 \\
        ResBlock   & 4$\times$4$\times$512 \\
        Conv2d 3$\times$3 & 4$\times$4$\times$512 \\
        Conv2d 3$\times$3 & 4$\times$4$\times$128 \\
      \end{tabular}
      \caption{$\xi^{\text{content}}$ architecture}
    \end{minipage}
    \begin{minipage}{.33\linewidth}
      \centering
      \begin{tabular}{c|c}
        \textbf{Layer} & \textbf{Output dimension}\\
        \hline
        Conv2d 3$\times$3 & 32$\times$32$\times$512 \\
        AvgPool2d 32$\times$32   & 512 \\
        Linear   & 256 \\
        \\\\

      \end{tabular}
      \caption{$\xi^{\text{theme}}$ architecture}
    \end{minipage}

  \end{center}
\end{table}

The above tables show the architecture for each component.
$\xi^{\text{feat}}$ takes $x$ as input and consists of several convolution layers whose output is passed to the two heads.
Conv2d 3$\times$3 denotes a 2D convolution layer with 3$\times$3 filters and padding of 1 to produce the same spatial dimension as input.
ResBlock denotes a residual block~\cite{he2016deep} with downsampling by 2$\times$ which is composed of two 3$\times$3 convolution layers and a skip connection layer.
After each layer, we put the leaky ReLU~\cite{maas2013rectifier} activation function, except for the last layer of $\xi^{\text{content}}$ and $\xi^{\text{theme}}$.
The outputs of $\xi^{\text{content}}$ and $\xi^{\text{theme}}$ are equally split into two chunks by the channel dimension, and used as $\mu$ and $\sigma$ for the reparameterization steps:
\begin{equation}
\label{eq:reparam}
    z = \mu + \epsilon \sigma, \quad \epsilon \sim N(0, I)
\end{equation}
producing $z^{\text{theme}} \in \mathbb{R}^{128}$ and $z^{\text{content}} \in \mathbb{R}^{4\times 4\times 64}$.

\subsubsection{Generator}
The generator architecture closely follows the generator of StyleGAN~\cite{karras2020analyzing}.
Here, we discuss a few differences.
$z^{\text{content}}$ goes through a 3$\times$3 convolution layer to make it a 4$\times$4$\times$512 tensor.
StyleGAN takes a constant tensor as an input to the first layer.
We concatenate the constant tensor with $z^{\text{content}}$ channel-wise and pass it to the first layer.
$z^{\text{theme}}$ goes through 8 linear layers, each outputting a 1024-dimensional vector, and the output is used for the adaptive instance normalization layers in the same way \textit{style} vectors are used in StyleGAN.
The generator outputs a 256$\times$256$\times$3 image.

\subsubsection{Discriminator}
Dicriminator takes the real and generated images (256$\times$256$\times$3) as input.
We use multi-scale multi-patch discriminators~\cite{vid2vid,isola2017image,shaham2019singan}, which results in higher quality images for complex scenes.

\begin{table}[h!]
  \begin{center}
    \label{tab:pretrain_disc}
    \begin{minipage}{.33\linewidth}
        \centering
        \begin{tabular}{c|c}
          \textbf{Layer} & \textbf{Output dimension}\\
          \hline
          Conv2d 3$\times$3 & 256$\times$256$\times$128 \\
          ResBlock   & 128$\times$128$\times$256 \\
          ResBlock   & 64$\times$64$\times$512 \\
          ResBlock   & 32$\times$32$\times$512 \\
          ResBlock   & 16$\times$16$\times$156 \\
          ResBlock   & 8$\times$8$\times$512 \\
          ResBlock   & 4$\times$4$\times$512 \\
          Conv2d 3$\times$3    & 4$\times$4$\times$512 \\
          Linear & 512 \\
          Linear & 1 \\

        \end{tabular}
        \caption{$D_1$ architecture}
    \end{minipage}%
    \begin{minipage}{.3\linewidth}
      \centering
      \begin{tabular}{c|c}
        \textbf{Layer} & \textbf{Output dimension}\\
        \hline
        Conv2d 3$\times$3 & 256$\times$256$\times$128 \\
        ResBlock   & 128$\times$128$\times$256 \\
        ResBlock   & 64$\times$64$\times$512 \\
        ResBlock   & 32$\times$32$\times$512 \\
        ResBlock   & 16$\times$16$\times$512 \\
        Conv2d 3$\times$3    & 16$\times$16$\times$1 \\
        \\\\\\\\
      \end{tabular}
      \caption{$D_2$ architecture}
    \end{minipage}
    \begin{minipage}{.35\linewidth}
      \centering
      \begin{tabular}{c|c}
        \textbf{Layer} & \textbf{Output dimension}\\
        \hline
        Conv2d 3$\times$3 & 128$\times$128$\times$128 \\
        ResBlock   & 64$\times$64$\times$256 \\
        ResBlock   & 32$\times$32$\times$512 \\
        ResBlock   & 16$\times$16$\times$512 \\
        ResBlock   & 8$\times$8$\times$512 \\
        Conv2d 3$\times$3    & 8$\times$8$\times$1 \\
        \\\\\\\\

      \end{tabular}
      \caption{$D_3$ architecture}
    \end{minipage}

  \end{center}
\end{table}

We use three discrminators $D_1, D_2,$ and $D_3$.
$D_1$ takes a 256$\times$256$\times$3 image as input and produces a single number.
$D_2$ takes a 256$\times$256$\times$3 image as input and produces 16$\times$16 patches each with a single number.
$D_3$ takes a 128$\times$128$\times$3 image as input and produces 8$\times$8 patches each with a single number.
The adversarial losses for $D_2$ and $D_3$ are averaged across the patches.
The inputs to $D_1,D_2,D_3$ are the real and generated images, except that the input to $D_3$ is downsampled by 2$\times$.
The model architectures are described in the above tables, and we use the same convolution layer and residual blocks from the previous sections.
Each layer is followed by a leaky ReLU activation function except for the last layer.

\subsubsection{Training}
We combine the loss functions of VAE~\cite{kingma2014autoencoding} and GAN~\cite{gan}, and let $L_{pretrain} = L_{VAE} + L_{GAN}$.
We use the same loss function for the adversarial loss $L_{GAN}$ from StyleGAN~\cite{karras2020analyzing}, except that we have three terms for each discriminator.
$L_{VAE}$ is defined as:
\begin{equation*}
\label{eqn:vae}
L_{VAE} = E_{z\sim q(z|x)}[log(p(x|z))] + \beta KL(q(z|x) || p(z))
\end{equation*}
where $p(z)$ is the standard normal prior distribution, $q(z|x)$ is the approximate posterior from the encoder $\xi$,
and $KL$ is the Kullback-Leibler divergence.
For the reconstruction term, we reduce the perceptual distance~\cite{zhang2018unreasonable} between the input and output images rather than the pixel-wise distance, and this term is weighted by 25.0.
We use separate $\beta$ values $\beta^{\text{theme}}$ and $\beta^{\text{content}}$ for $z^{\text{content}}$ and $z^{\text{theme}}$.
We also found different $\beta$ values work better for different environments.
We use $\beta^{\text{theme}}=1.0, \beta^{\text{content}}=2.0$ for Carla,
$\beta^{\text{theme}}=1.0, \beta^{\text{content}}=4.0$ for Gibson, and
$\beta^{\text{theme}}=1.0, \beta^{\text{content}}=1.0$ for RWD.
Adam~\cite{kingma2014adam} optimizer is employed with learning rate of 0.002 for 310,000 optimization steps.
We use a batch size of 16.

\subsection{Dynamics Engine}
\label{sec:dynamics_engine}
With the pre-trained encoder and decoder, the Dynamics Engine learns the transition between latent codes from one time step to the next given an action $a_t$.
We first pre-extract the latent codes for each image in the training data, and only learn the transition between the latent codes.
All neural network layers described below are followed by a leaky ReLU activation function, except for the outputs of discriminators, the outputs for $\mu, \sigma$ variables used for reparameterization steps, and the outputs for the AdaIN parameters.

\begin{figure}[t!]
    \begin{center}
        \includegraphics[width=0.5\linewidth]{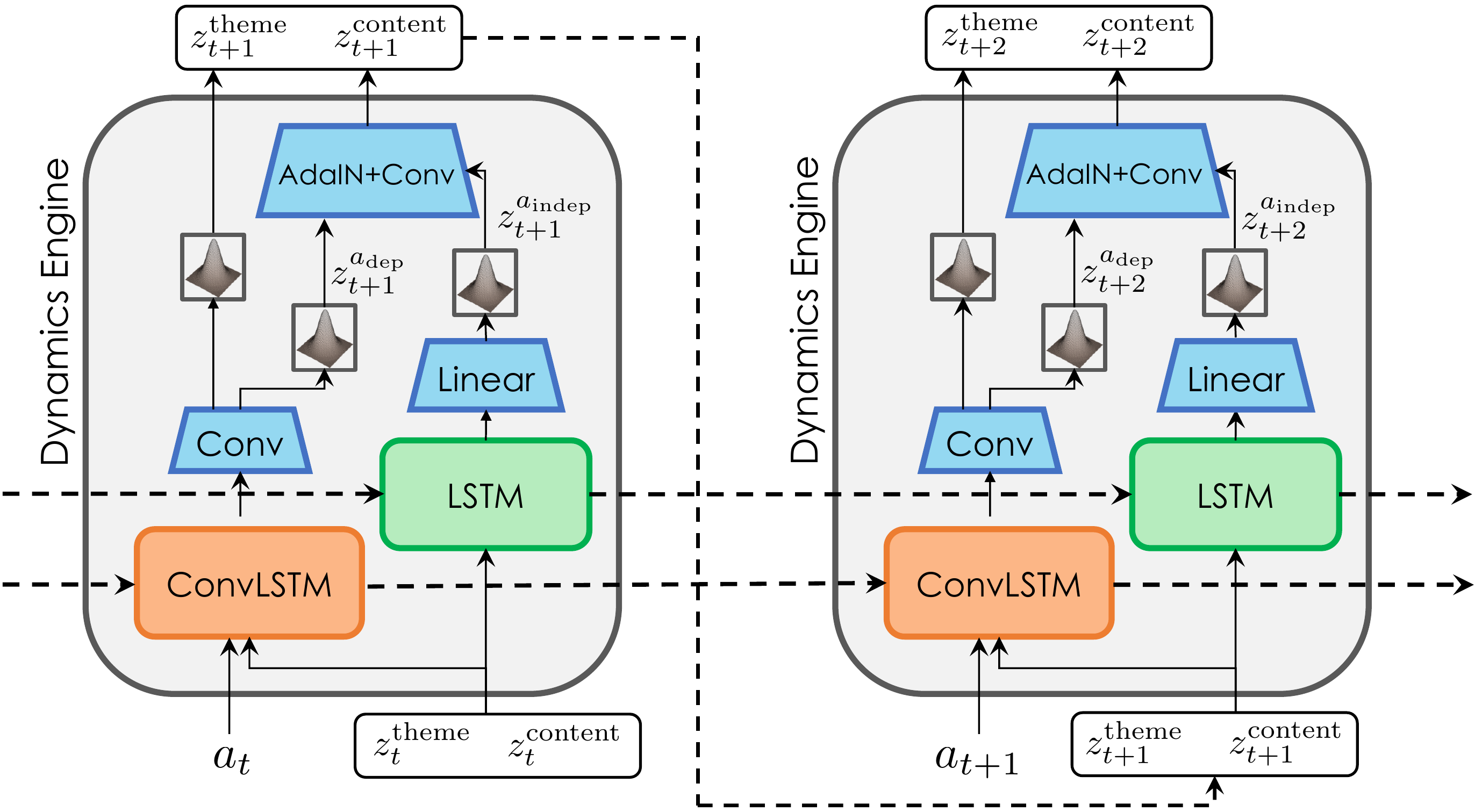}
    \end{center}
    \caption[]{
        \small Dynamics Engine produces the next latent codes, given an action and previous latent codes.
        It disentangles content information into action-dependent and action-independent features with its two separate LSTMs.
        Dashed lines correspond to temporal connections.
         Gaussian blocks indicate reparameterization steps. 
    }
    \label{fig:model_dynamics_engine}
\end{figure}

The major components of the Dynamics Engine are its two LSTM modules.
The first one learns the spatial transition between the latent codes and is implemented as a convolutional LSTM module (Figure~\ref{fig:model_dynamics_engine}).
\begin{equation}
    v_t = \mathcal{F}(\mathcal{H}(h^{\text{conv}}_{t-1}, a_t, z_t^{\text{content}}, z_t^{\text{theme}}))
\end{equation}
\begin{equation}
    i_t, f_t, o_t = \sigma(v_t^i),\sigma(v_t^f), \sigma(v_t^o)
\end{equation}
\begin{equation}
    c^{\text{conv}}_{t} = f_t \odot c^{\text{conv}}_{t-1} + i_t \odot \tanh(v_t^g)
\end{equation}
\begin{equation}
    h^{\text{conv}}_{t} = o_t \odot \tanh(c^{\text{conv}}_{t})
\end{equation}
where $h^{\text{conv}}_{t}, c^{\text{conv}}_{t}$ are the hidden and cell state of the convLSTM module, and $i_t, f_t, o_t$ are the input, forget, output gates, respectively.
$\mathcal{H}$ replicates $a_t$ and $z_t^{\text{theme}}$ spatially to match the $4\times 4$ spatial dimension of $z^{\text{content}}_t$.
It fuses all inputs by concatenating and running through a $1\times 1$ convolution layer, resulting in a 4$\times$4$\times$48 tensor.
$\mathcal{F}$ is composed of two $3\times 3$ convolution layers with a padding of 1, and produces $v_t \in \mathbb{R}^{4\times 4\times 512}$.
$v_t$ is split channel-wise into intermediate variables $v_t^i, v_t^f, v_t^o, v_t^g$.
All state and intermediate variables have the same size $\mathbb{R}^{4\times 4\times 128}$.
The hidden state $h^{\text{conv}}_{t}$ goes through two separate convolution layers: 1) 1$\times$1 Conv2d layer that produces 4$\times$4$\times$128 tensor which is split into two chuncks with equal size 4$\times$4$\times$64 and used for the reparameterization step (Eq.~\ref{eq:reparam}) to produce $z_{t+1}^{a_{\text{dep}}} \in \mathbb{R}^{4\times 4\times 64}$,
and 2) 4$\times$4 conv2d layer with no padding that produces a 256 dimensional vector; this is also split into two chunks and reparameterized to produce $z_{t+1}^{\text{theme}} \in \mathbb{R}^{128}$.

The second one is a plain LSTM~\cite{hochreiter1997long} module that only takes $z_t$ as input.
Therefore, this module is responsible for information that does not depend on the action $a_t$.
The input $z_t$ is flattened into a vector $\in \mathbb{R}^{1152}$ and goes through five linear layers each outputting 1024-dimensional vectors.
The encoded $z_t$ is fed to the LSTM module and all variables inside this module have size $\mathbb{R}^{1024}$.
We experimented with both LSTM and GRU~\cite{chung2014empirical} but did not observe much difference.
The hidden state goes through a linear layer that outputs a 2048-dimensional vector.
This vector is split into two chunks for reparmetrization and produces $z_{t+1}^{a_{\text{indep}}} \in \mathbb{R}^{1024}$.

Finally, $z_{t+1}^{a_{\text{dep}}}$ and $z_{t+1}^{a_{\text{indep}}}$ are used as inputs to two $AdaIN$ + Conv blocks.
\begin{equation}
    \bm{\alpha}, \bm{\beta} = MLP(z_{t+1}^{a_{\text{indep}}})
\end{equation}
\begin{equation}
    z_{t+1}^{\text{content}} = \mathcal{C}(\mathcal{A}(\mathcal{C}(\mathcal{A}(z_{t+1}^{a_{\text{dep}}}, \bm{\alpha}, \bm{\beta})),\bm{\alpha}, \bm{\beta}))
\end{equation}
where we denote convolution and $AdaIN$ layers as $\mathcal{C}$ and $\mathcal{A}$, respectively.
The two $MLP$s (for each block) consist of two linear layers.
They produce 64 and 256 dimensional $\bm{\alpha}, \bm{\beta}$, respectively.
The first 3$\times$3 conv2d layer $\mathcal{C}$ produces 4$\times$4$\times$256 tensor, and the second 3$\times$3 conv2d layer produces $z_{t+1}^{\text{content}} \in \mathbb{R}^{4\times 4\times 64}$.

\subsubsection{Discriminator}
We use disciminators on the flattened 1152 dimensional latent codes $z$ (concatenation of $z^{\text{theme}}$ and flattened $z^{\text{content}}$).
There are two discriminators 1) single latent discriminator $D_{single}$, and 2) temporal action-conditioned discriminator $D_{temporal}$.

\begin{table}[h!]
  \begin{center}
    \label{tab:pretrain_disc}
    \begin{minipage}{.35\linewidth}
        \centering
        \begin{tabular}{c|c}
          \textbf{Layer} & \textbf{Output dimension}\\
          \hline
          SNLinear + BN & 1024 \\
          SNLinear + BN & 1024 \\
          SNLinear + BN & 1024 \\
          SNLinear + BN & 1024 \\
          SNLinear + BN & 1024 \\
          SNLinear & 1 \\
        \end{tabular}
        \caption{$D_{single}$ architecture}
    \end{minipage}%
    \begin{minipage}{.6\linewidth}
        \label{tab:temporal_disc}
        \centering
        \begin{tabular}{c|c| c}
          \textbf{Layer} & \textbf{Input dimension} & \textbf{Output dimension}\\
          \hline
          SNConv1d & 2048$\times$31 & 128$\times$15 \\
          SNConv1d & 128$\times$15 & 256$\times$13 \\
          SNConv1d & 256$\times$13 & 512$\times$6 \\
        \end{tabular}
        \caption{
            $D_{temporal}$ architecture.
            Input and output dimensions contain two numbers, the first one for the number of channels or vector dimension, and the second one for the temporal dimension.
            Note that Conv1d is applied on the temporal dimension.
        }
    \end{minipage}%

  \end{center}
\end{table}

We denote SNLinear and SNConv as linear and convolution layers with Spectral Normalization~\cite{miyato2018spectral} applied, and BN as 1D Batch Normalization layers~\cite{ioffe2015batch}.
$D_{single}$ is a 6-layer $MLP$ that tries to discriminate generated $z$ from the real latent codes.
It takes a single $z$ as input and produces a single number.
For the temporal action-conditioned discriminator $D_{temporal}$, we first reuse the 1024-dimensional feature representation from the fourth layer of $D_{single}$ for each $z_t$.
The represenations for $z_t$ and $z_{t-1}$ are concatenated and go through a SNLinear layer to produce the 1024-dimensional temporal discriminator feature.
Let us denote the temporal discriminator feature as $z_{t,t-1}$.
The action $a_t$ also goes through a SNLinear layer to produce the 1024-dimensional action embedding.
$z_{t,t-1}$ and the action embedding are concatenated and used as the input to $D_{temporal}$.
We use 32 time-steps to train DriveGAN, so the input to $D_{temporal}$ has size 2048$\times$31 where 31 is the temporal dimension.
Table 8 shows the architecture of $D_{temporal}$.
After each layer of $D_{temporal}$, we put a 3-timestep wide convolution layer that produces a single number for each resulting time dimension. Therefore, there are three outputs of $D_{temporal}$ with sizes 14, 11, and 4 which can be thought of as \textit{patches} in the temporal dimension.
We also sample negative actions $\bar{a}_t$, and the job of $D_{temporal}$ is to figure out if the given sequence of latent codes is realistic and faithful to the given action sequences.
$\bar{a}_t$ is sampled randomly from the training dataset.

\subsubsection{Training}
We use Adam optimizer with learning rate of 0.0001 for 400,000 optimization steps.
We use batch size of 128 each with 32 time-steps and train with a warm-up phase.
In the warm-up phase, we feed in the ground-truth latent codes as input for the first 18 time-steps and linearly decay the number to 1 at 100-th epoch, which corresponds to completely autoregressive training at that point.
We use the loss $L_{DE} = L_{adv} + L_{latent} + L_{action} + L_{KL}$.
$L_{adv}$ is the adversarial losses, and we use the hinge loss~\cite{lim2017geometric, tran2017hierarchical}. We also add a $R_1$ gradient regularizer~\cite{mescheder2018training} to $L_{adv}$ that penalizes the gradients of discriminators on true data .
$L_{action}$ is the action reconstruction loss (implemented as a mean squared error loss) which we obtain by running the temporal discriminator features $z_{t,t-1}$ through a linear layer to reconstruct the input action $a_{t-1}$.
Finally, we add the latent code reconstruction loss $L_{latent}$ (implemented as a mean squared error loss) so that the generated $z_t$ matches the input latent codes, and reduce the $KL$ penalty $L_{KL}$ for $z_t^{a_{\text{dep}}}$,$z_t^{a_{\text{indep}}}$, $z_t^{\text{theme}}$.
$L_{latent}$ is weighted by 10.0 and we use different $\beta$ for the $KL$ penalty terms.
We use $\beta^{a_{\text{dep}}}=0.1, \beta^{a_{\text{indep}}}=0.1, \beta^{\text{theme}}=1.0$ for Carla, and
$\beta^{a_{\text{dep}}}=0.5, \beta^{a_{\text{indep}}}=0.25, \beta^{\text{theme}}=1.0$ for Gibson and RWD.

\section{Additional Analysis on Experiments}

\textbf{Multi-patch Multi-scale discriminator}
We experimented with Carla dataset to choose the image discriminator architecture.
In contrast to the plain StyleGAN, the datasets studied in this work contain much more diverse objects in multiple locations.
Using a multi-patch multi-scale discriminator~\cite{vid2vid,isola2017image,shaham2019singan} improved our FID score on Carla images from \textbf{72.3} to \textbf{67.1} over the StyleGAN discriminator.

\textbf{LiftSplat}
\cite{philion2020lift} proposed a model for producing the Bird's-Eye-View (BEV) representation of a scene from camera images.
Section 4.3 in the main text shows how we can leverage LiftSplat to get BEV lane predictions from a simulated sequence from DriveGAN.
We can further analyze the qualitative result by comparing how the perception model (LiftSplat) perceives the ground truth and generated sequences differently.
We fit a quadratic function to the LiftSplat BEV lane prediction for each image in the ground-truth sequence, and compare the distance between the fitted quadratic and the predicted lanes.

\begin{table}[h!]
  \begin{center}
    \label{tab:pretrain_disc}
    \begin{minipage}{.7\linewidth}
        \centering
        \begin{tabular}{c|c c c c}
           & \multicolumn{4}{c}{\textbf{BEV Prediction Look-ahead Distance}}\\
          \textbf{Model} &     25m & 50m & 75m & 100m \\
          \hline
          Random         & 0.91m & 1.78m & 2.95m & 4.74m \\
          DriveGAN       & 0.58m & 1.00m & 1.70m & 2.99m \\
          Ground-Truth   & 0.31m & 0.37m & 0.88m & 2.07m \\
        \end{tabular}
        \caption{
        Mean distance from the BEV lane predictions and the fitted quadratic function in meters.
        }
    \end{minipage}
  \end{center}
\end{table}

We show results on different look-ahead distances, which denote how far from the ego-car we are making the BEV predictions for.
The above table lists the mean distance from the BEV lane predictions and the fitted quadratic function.
\textit{Random} compares the distance between the fitted quadratic and the BEV prediction for a randomly sampled RWD sequence.
\textit{DriveGAN} compares the distance for the BEV prediction for the optimized sequence with \textit{differentiable simulation} of DriveGAN.
\textit{Ground-Truth} compares the distance for the BEV prediction for the ground-truth image.
Note that \textit{Ground-Truth} is not 0 since the fitted quadratic does not necessarily follow the lane prediction from the ground-truth image exactly.
We can see that DriveGAN-optimized sequences produce lanes that follow the ground-truth lanes, which demonstrates how we could find the underlying actions and stochastic variables from a real video through differentiable simulation.

\section{DriveGAN Simulator User Interface}
\begin{figure}[h!]
    \begin{center}
        \includegraphics[width=0.75\linewidth]{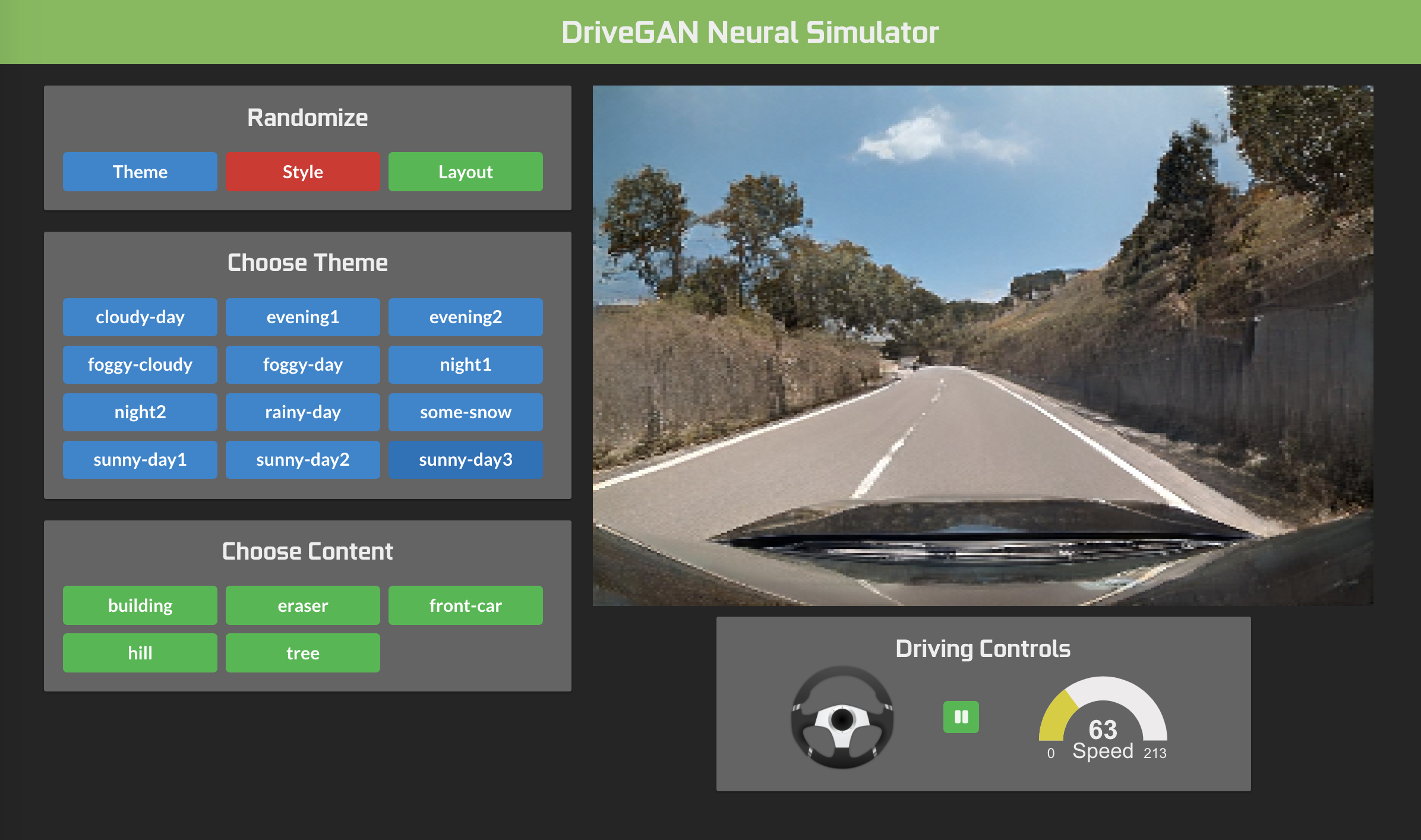}
    \end{center}
    \caption[]{
        \small UI for DriveGAN simulator
    }
    \label{fig:driveGAN_demo}
\end{figure}

We build an interactive user interface for users to play with DriveGAN.
Figure~\ref{fig:driveGAN_demo} shows the application screen.
It has controls for the steering wheel and speed, which can be controlled by the keyboard.
We can randomize different components by sampling $z^{a_{\text{indep}}}, z^{\text{content}}$ or $z^{\text{theme}}$.
We also provide a pre-defined list of themes and objects that users can selectively use for specific changes.
The supplementary video demonstrates how this UI can enable interactive simulation.



\end{document}